\documentclass[journal]{IEEEtran}
\usepackage{amsmath,amsfonts}
\usepackage{array}
\usepackage[caption=false,font=normalsize,labelfont=sf,textfont=sf]{subfig}
\usepackage{textcomp}
\usepackage{stfloats}
\usepackage{url}
\usepackage{verbatim}
\usepackage{graphicx}
\usepackage{cite}
\usepackage[colorlinks=true, linkcolor=blue, citecolor=blue, urlcolor=blue]{hyperref}

\usepackage[table]{xcolor}
\usepackage{colortbl}

\usepackage{times}
\usepackage{soul}
\usepackage{url}
\usepackage[utf8]{inputenc}
\usepackage{graphicx}
\usepackage{amsmath}
\usepackage{amsthm}
\usepackage{booktabs}
\usepackage{algorithm}
\usepackage{algorithmicx}
\usepackage[switch]{lineno}
\usepackage{quoting}
\usepackage{multirow}
\usepackage{xcolor}
\usepackage{amssymb}
\usepackage{amssymb}
\usepackage{algpseudocode}
\newcommand{\ours}{AVS-Mamba}

\begin{document}

\title{AVS-Mamba: Exploring Temporal and Multi-modal Mamba for Audio-Visual Segmentation}

\author{Sitong Gong, Yunzhi Zhuge,~\IEEEmembership{Member,~IEEE}, Lu Zhang, Yifan Wang,~\IEEEmembership{Member,~IEEE}, Pingping Zhang,~\IEEEmembership{Member,~IEEE}, Lijun Wang,~\IEEEmembership{Member,~IEEE}, Huchuan Lu,~\IEEEmembership{Fellow,~IEEE}

        % <-this % stops a space
\thanks{Manuscript received 2 January 2024; revised 23 April 2024; accepted
20 June 2024. \textit{(Corresponding author: Yunzhi Zhuge.)}}
\thanks{Sitong Gong, Yunzhi Zhuge and Lu Zhang are with the School of Information and Communication Engineering, Dalian University of Technology, Dalian 116081, China. (e-mail: stgong@mail.dlut.edu.cn; zgyz@dlut.edu.cn; zhangluu@dlut.edu.cn)}
\thanks{Yifan Wang is with the School of Innovation and Entrepreneurship, Dalian University of Technology, Dalian 116081, China. (email: wyfan@dlut.edu.cn)}
\thanks{Pingping Zhang Lijun Wang and Huchuan Lu are with the School of Future Technology and the School of Artificial Intelligence, Dalian University of Technology, Dalian 116081, China. (email: zhpp@@dlut.edu.cn; ljwang@dlut.edu.cn; lhchuan@dlut.edu.cn)}}

\markboth{IEEE Transactions on Multimedia}%
{Shell \MakeLowercase{\textit{et al.}}: A Sample Article Using IEEEtran.cls for IEEE Journals}


\maketitle

\begin{abstract}
The essence of audio-visual segmentation (AVS) lies in locating and delineating sound-emitting objects within a video stream. While Transformer-based methods have shown promise, their handling of long-range dependencies struggles due to quadratic computational costs, presenting a bottleneck in complex scenarios. To overcome this limitation and facilitate complex multi-modal comprehension with linear complexity, we introduce AVS-Mamba, a selective state space model to address the AVS task. Our framework incorporates two key components for video understanding and cross-modal learning: Temporal Mamba Block for sequential video processing and Vision-to-Audio Fusion Block for advanced audio-vision integration. Building on this, we develop the Multi-scale Temporal Encoder, aimed at enhancing the learning of visual features across scales, facilitating the perception of intra- and inter-frame information. To perform multi-modal fusion, we propose the Modality Aggregation Decoder, leveraging the Vision-to-Audio Fusion Block to integrate visual features into audio features across both frame and temporal levels.
Further, we adopt the Contextual Integration Pyramid to perform audio-to-vision spatial-temporal context collaboration. Through these innovative contributions, our approach achieves new state-of-the-art results on the AVSBench-object and AVSBench-semantic datasets. Our source code and model weights are available at \href{https://github.com/SitongGong/AVS-Mamba}{AVS-Mamba}.
\end{abstract}


\section{Introduction}
\begin{figure}
    \centering
    \vspace{-2mm}
    \includegraphics[width=1\linewidth]{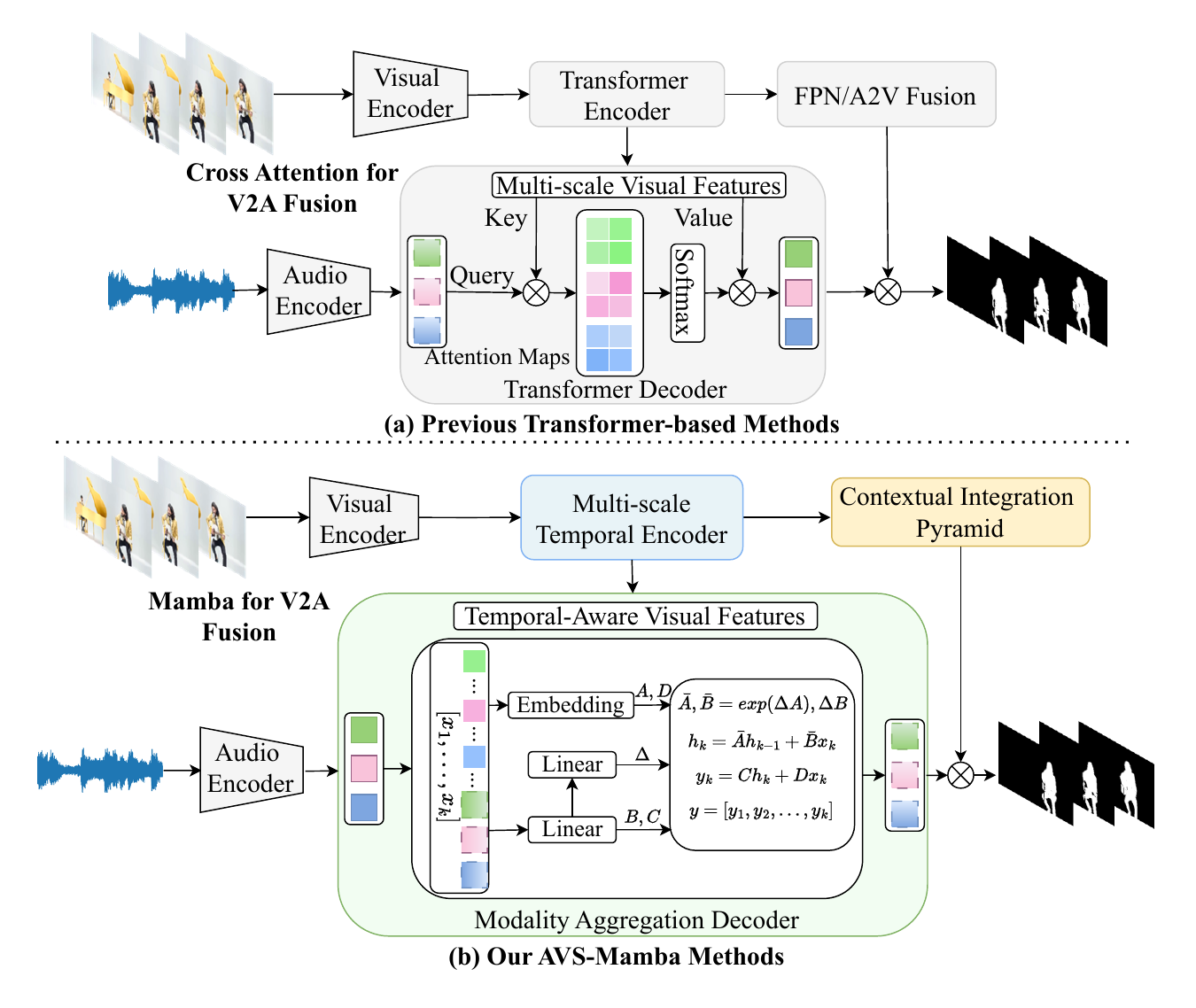}
    \vspace{-7mm}
    \caption{Comparison with previous Transformer-based methods. (a) Previous methods use cross-attention in the Decoder to query visual features with audio features. (b) Our AVS-Mamba serializes visual features into 1D sequences and merges them with audio features through selective scanning mechanism.
    % , enabling efficient cross-modal feature integration via state compression and dynamic parameterization.
    }
    \vspace{-4mm}
    \label{fig:1}
\end{figure}

% 第一段先介绍一下AVS任务，而后分开介绍三个子任务面临的问题。
\IEEEPARstart{H}{uman} perception and understanding of the environment are accomplished through the simultaneous integration of visual and auditory inputs. Auditory information plays a crucial role in enhancing the visual system's ability to localize referential objects, thereby facilitating subsequent actions. Consequently, the task of audio-visual segmentation (AVS)~\cite{zhou2022audio} has emerged, focusing on the identification and segmentation of sounding objects within video frames under the guidance of audio cues, reflecting a concerted effort to mirror the intricacies of human sensory integration.
The audio-visual segmentation task encompasses three sub-tasks, each presenting challenges across multiple dimensions. To begin with, when only a single sound source persists throughout the entire video, it is necessary to perform profound multi-modal integration and alignment, establishing an explicit association between auditory signals and their visual counterparts. Then, in more complex scenarios involving multiple sound sources, effectively capturing and utilizing temporal audio cues alongside inter-frame visual relationships becomes essential to minimize frame-to-frame interference. Finally, for audio-visual semantic segmentation (AVSS)~\cite{zhou2023audio}, the task demands the extraction and adaptation of semantic content from auditory inputs into the visual domain.

% The AVS encompasses three sub-tasks with challenges spanning multiple dimensions: 
% (1) While only a single sound source persists throughout the entire video, it is necessary to perform profound multi-modal integration and alignment, building an explicit association between auditory signals and their visual counterparts.  
% (2) In complex scenarios with multiple sound sources, effectively capturing and utilizing temporal audio cues alongside inter-frame visual relationships is essential for minimizing frame-to-frame interference.
% (3) As for the audio-visual semantic segmentation~\cite{zhou2023audio} (AVSS), it demands the extraction and adaptation of semantic content from auditory inputs into the visual domain. 

% 第二段概括一下之前的AVS方法及其存在的问题，其中基于Transformer的方法要从促进特征融合的角度来写。
The design of existing AVS methods~\cite{zhou2022audio, hao2023improving, gao2023avsegformer, huang2023discovering, li2023catr, chen2024bootstrapping} primarily relies on the Convolutional Neural Networks (CNNs) and Transformer.  
Earlier methods~\cite{zhou2022audio, hao2023improving} utilize CNNs for local feature extraction and dense multi-modal fusion, falling short in capturing broader visual context due to CNNs' limited local receptive fields. 
In contrast, Transformer-based approaches~\cite{gao2023avsegformer, huang2023discovering, li2023catr, chen2024bootstrapping} leverage attention mechanisms for global audio-vision feature integration. These methods employ audio query guidance to access and focus on the corresponding visual contextual information, thereby achieving more accurate segmentation results.
Despite the adoption of Transformer enhances comprehensive feature interactions, these methods still face significant issues. Firstly, the quadratic complexity of Transformers limits the number of tokens that can be processed concurrently, which contradicts the model’s ability to efficiently manage multi-scale long video sequences. Additionally, previous methods have primarily focused on the interaction of multi-scale visual features within individual frames, thereby neglecting crucial fusion and perception across frames. For the cross-modal decoder, earlier approaches have utilized audio queries to focus on the features across all frames in the temporal dimension but have failed to consider the strong correlations between adjacent frames. This oversight results in an underutilization of spatiotemporal consistency within the multi-modal features.

\begin{figure*}
    \centering
    \vspace{-2mm}
    \includegraphics[width=1.03\linewidth]{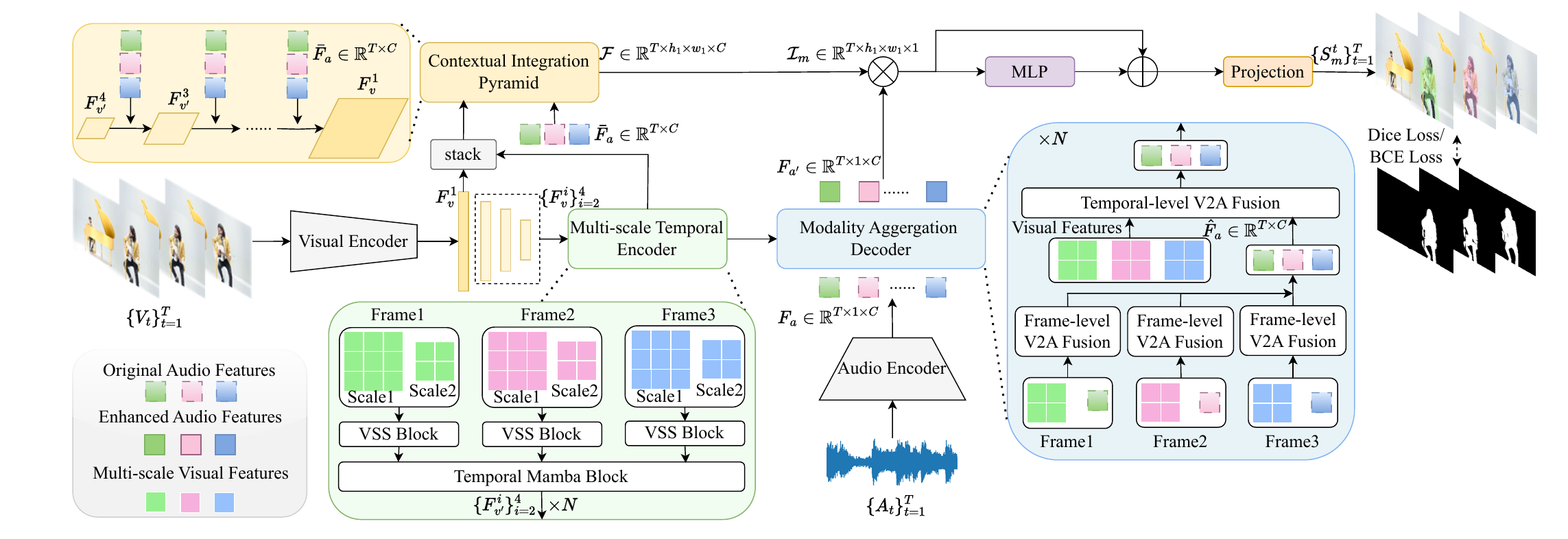}
    \vspace{-5mm}
    \caption{The overall architecture of our proposed AVS-Mamba. The Multi-scale Temporal Encoder, incorporating VMamba and Temporal Mamba structures, processes spatial and temporal associations of multi-scale visual features. The Modality Aggregation Decoder utilizes the V2A Fusion Block for multi-directional serialization and integration of multi-modal features, enhancing information transfer from visual to audio modalities. Finally, the Contextual Integration Pyramid, integrating the FPN structure with refined Mamba modules, facilitates deep spatial-temporal interactions and cross-scale feature fusion.}
    
    % Our model begins by delineating the spatial and temporal associations of multi-scale visual features, facilitated by the VMamba and Temporal Mamba structures within the Multi-scale Temporal Encoder. The Modality Aggregation Decoder employs the V2A Fusion block for multi-directional serialization of multi-modal features, culminating in a comprehensive information gathering from visual to audio features. The Contextual Integration Pyramid achieves in-depth spatial-temporal interactions and cross-scale feature fusion by harmonizing the FPN structure with our refined Mamba modules.}
    \label{fig:2}
    \vspace{-1mm}
\end{figure*}

% 这里重点介绍一下Mamba的应用以及优势，因为上次被质疑缺少动机阐述
The introduction of the Selective State Space Model (Mamba)~\cite{gu2023mamba} 
significantly influenced research due to its ability to efficiently model long-range interactions with linear computational complexity. Leveraging techniques such as selective scanning and dynamic weighting, Mamba has demonstrated exceptional performance in language sequence modeling and has been effectively adapted to visual tasks\cite{zhu2024vision, liu2024vmamba}, showcasing its versatility across different domains.
% Motivated by Mamba's effectiveness in language sequence modeling, researchers are increasingly exploring its application across various vision tasks~\cite{zhu2024vision, liu2024vmamba}.
We have identified several intrinsic advantages of Mamba that could significantly benefit AVS tasks: Firstly, its ability to process longer sequences in a single pass enables the exchange of multi-frame video features across various scales. Additionally, by modeling sequences in causal order, Mamba strengthens the connections between adjacent frames while effectively minimizing interference from redundant information in temporally distant frames.
Building on the insights from our analysis, we present \ours, which capitalizes on the Mamba framework to facilitate efficient cross-frame sequence modeling and enhance cross-modal feature interactions, thereby optimizing audio-visual segmentation tasks.
As depicted in Fig.~\ref{fig:1}, in contrast to prior methods that utilize cross-attention mechanisms for fusing and aligning visual and auditory modalities, AVS-Mamba innovatively exploits Mamba's state compression mechanism and dynamic parameterization to address AVS.
The overall architecture of the proposed method is shown in Fig.~\ref{fig:2}.
To begin with, the Multi-scale Temporal Encoder (MTE) is introduced to enhance cross-scale feature interactions within frames through the Visual State Space (VSS) Block and to facilitate cross-frame connectivity using the Temporal Mamba Block.
Subsequently, the Modality Aggregation Decoder (MAD) receives multi-scale visual and audio features as input for bi-modal processing. 
Differing from the Transformer decoder that associates audio queries with visual features from the corresponding frame via cross-attention, MAD leverages the Vision-to-Audio Fusion Block for frame- and temporal-level integration, promoting the awareness of inter-frame relations. 
% This process begins with serializing visual features according to varied scanning sequences, followed by the incorporation of audio features into the sequence. 
Furthermore, we develop a Contextual Integration Pyramid (CIP), combining our specially designed Mamba-based modules with the Feature Pyramid Network (FPN)~\cite{lin2017feature}. Specifically, visual features across different scales initially undergo cross-frame interactions, which are then fused with auditory information. The mask features are augmented by progressive upsampling and integrative processing, drawing on multi-scale and multi-modality enhancements.
% Accompanied by progressive upsampling and integrative processes, the mask features are augmented, drawing on multi-scale and multi-modality enhancements.

Our contributions can be summarized as follows: 
\vspace{-1mm}
\begin{itemize}
\item We present \ours, a novel architecture that leverages Mamba's potential for efficiently capturing long-range dependencies in audio-visual segmentation while mitigating its sensitivity to scanning direction. The architecture incorporates cross-scale interactions and a temporal-aware design strategy, offering significant insights for extending Mamba's application to diverse multimodal learning tasks.
% To our knowledge, this is the first time that Mamba is applied to Audio-Visual Learning, encouraging further investigation into its utility in multi-modal perception.

\item We propose several task-specific modules to enhance performance. The MTE module is designed to enhance cross-scale temporal relationships, strengthening connectivity between consecutive frames. The MAD module effectively aggregates multi-directional vision-to-audio features spanning spatial-temporal dimensions. The CIP module executes intensive cross-frame audio-to-vision interactions to produce mask features.

\item Extensive experiments demonstrate that AVS-Mamba establishes new state-of-the-art benchmarks on the AVSBench-object and AVSBench-semantic datasets, excelling in terms of both the Jaccard index and F-score metrics, underscoring the effectiveness of the proposed approach.
\end{itemize}

\section{Related Works}

\subsection{Audio-Visual Segmentation}
Audio-visual segmentation (AVS) stands apart from other audio-visual representation learning tasks like audio-visual speech enhancement~\cite{xiong2022look, wang2020robust}, audio-visual event localization~\cite{xue2021audio, jiang2023leveraging, liu2022dense, feng2023css}, and sound source localization~\cite{liu2022visual, mo2023audio, fu2023multimodal}. Unlike these tasks, which primarily localize audio-visual content, AVS~\cite{zhou2022audio, zhou2023audio} aims for pixel-level segmentation of sound-emitting objects using audio cues. The key method in AVS, TPAVI~\cite{zhou2022audio}, leverages dense interactions between visual and audio features via CNNs but faces limitations due to locality, resulting in redundant information in two-dimensional space.

Inspired by advances in video transformers~\cite{xian2023vita, li2024adaptive, zhang2023end,zhuge2024learning}, subsequent approaches utilizing Transformer-based architectures~\cite{gao2023avsegformer, huang2023discovering, li2023catr, chen2024bootstrapping} have refined the process of integrating features across various scales and modalities, achieving significant performance improvements. For instance, AVSegFormer~\cite{gao2023avsegformer} leverages audio features to query multi-scale visual features within frames, enhancing the resolution of details about sounding objects. AQFormer~\cite{huang2023discovering} utilizes queries to assess visual features across frames, promoting the joint extraction of local and global relationships. CATR~\cite{li2023catr} advances bi-modal feature fusion by combining multiple modules, and AVSAC~\cite{chen2024bootstrapping} introduces the BAVD module, which strengthens audio cues while maintaining balance in cross-modal interactions. 
% Additionally, other methods explore AVS challenges through innovative frameworks such as latent diffusion models and variational auto-encoder architectures~\cite{mao2023contrastive, mao2023multimodal}.

Distinct from the approaches mentioned above, we introduce a novel audio-visual segmentation technique that utilizes the Mamba model. We have tailored task-specific modifications to optimally harness Mamba’s capabilities for long-range feature extraction, which is critical for effective temporal modeling across various scales and modalities.

% applies task-specific adjustments to effectively utilize Mamba's capacity in long-range feature extraction for temporal modeling across scales and modalities. 
% Distinguishing from the aforementioned approaches, we introduce an AVS strategy rooted in Mamba, leveraging task-specific refinements to tap into Mamba's potential for cross-modal integration. 
% Mamba这部分还需要再加入一些参考文献

\subsection{Mamba in Vision Tasks}
The Selective State Space Model (Mamba) offers an innovative alternative to foundational models in vision, demonstrating superior capability in capturing long-range contexts with linearly increasing computational costs relative to sequence length, thus providing a more scalable solution. Notable visual models that incorporate Mamba technology include Vim~\cite{zhu2024vision} and VMamba~\cite{liu2024vmamba}. Vim builds upon Vision Transformers (ViTs)\cite{dosovitskiy2020image, touvron2021training} by integrating a bidirectional scanning strategy, allowing for enhanced processing of unfolded visual token sequences from both forward and backward directions. In contrast, VMamba employs the hierarchical architecture of Swin Transformers\cite{liu2021swin} with a 2D selective scan strategy, addressing challenges related to directional sensitivity.

A significant amount of research has applied Mamba to a diverse range of vision tasks. Notably, studies in medical image segmentation~\cite{yang2024vivim, yue2024medmamba, zhang2024vm, ye2024p} have utilized Mamba-based models to surpass existing benchmarks. In the field of remote sensing, researchers have harnessed Mamba's causal modeling capabilities, achieving considerable advances~\cite{ma2024rs, chen2024rsmamba, liu2024rscama, chen2024changemamba}. Mamba has also shown promise in low-level tasks~\cite{guo2024mambair, zheng2024u, shi2024vmambair}, and efforts have been made to enhance its proficiency in interpreting both image and linguistic sequences~\cite{qiao2024vlmamba, zhao2024cobra, yang2024remamber, liu2024robomamba}. Furthermore, the model has been adapted for video processing challenges~\cite{chen2024video, li2024videomamba, lu2024videomambapro}, time series forecasting~\cite{wang2024mamba} and infrared small target detection~\cite{chen2024mim}, with additional efforts focusing on refining the VMamba architecture to improve scanning sequences and computational efficiency~\cite{pei2024efficientvmamba, huang2024localmamba, yang2024plainmamba, shi2024multi}.

% Despite significant advancements in previous efforts, the potential of Mamba in audio-visual learning remains largely unexplored. 
During the development of our work, several concurrent works emerged. AVMamba~\cite{huangav} leveraged cross-modality selection mechanism within Mamba 
models to solve audio-visual question answering (AVQA)~\cite{li2022learning}, which served as the preliminary attempt of applying Mamba to audio-visual task. SelM~\cite{li2024selm} combined Mamba with Transformer modules to tackle the challenges in AVS. However, it remains limited to using Mamba for unimodal feature modeling and employs simple multiplication and attention mechanisms for coarse-grained fusion of audio-visual features.
Therefore, our research seeks to further unleash Mamba's capabilities to audio-visual segmentation tasks. 
% We specifically focus on improving segmentation performance by effectively integrating audio and visual modalities. 
Our goal is to fully leverage the strengths of the Mamba framework to enhance these complex multimodal interactions.

% Despite significant advancements in previous efforts, the potential of the Mamba framework in audio-visual segmentation remains unexplored. Therefore, our research seeks to extend Mamba's capabilities to audio-visual segmentation tasks. We are specifically focused on improving segmentation performance by effectively integrating audio and visual modalities. Our goal is to fully leverage the strengths of the Mamba framework to enhance these complex multimodal interactions.

\section{Preliminary}

\paragraph{State Space Models}
State Space Models (SSMs) constitute a category of models specifically designed for modeling one-dimensional sequences, drawing on principles from continuous linear time-invariant systems~\cite{kalman1960new}. These models transform an input $x(t)\in{\mathbb{R}^{L}}$ into an output $y(t)\in{\mathbb{R}^{L}}$ by learning a hidden state space $h(t)\in{\mathbb{R}^{N}}$, the process of which can be formulated as: 
\begin{equation}
\begin{aligned}
    h'(t)&=\mathbf{A}h(t) + \mathbf{B}x(t)\\
    y(t)&=\mathbf{C}h(t) + \mathbf{D}x(t)
\end{aligned} \label{con:ODE}
\end{equation}
where $\mathbf{A}\in{\mathbb{R}^{N\times{N}}}$ represents the evolution matrix, $\mathbf{B}\in{\mathbb{R}^{N}}$ and $\mathbf{C}\in{\mathbb{R}^{N}}$ are the projection matrix and $\mathbf{D}\in{\mathbb{R}^{1}}$ represents the skip connection. 

To incorporate State Space Models (SSMs) into deep learning contexts, the zero-order hold (ZOH) technique is employed to discretize the ordinary differential equations (ODEs) (Eq.~\ref{con:ODE}), with the discretization procedure detailed as: 
\begin{equation}
    \begin{aligned}
        \mathbf{\bar{A}}&=exp(\mathbf{\Delta{A}})\\
        \mathbf{\bar{B}}&=(exp(\mathbf{\Delta{A}})-I)(\mathbf{A})^{-1}\mathbf{B}\\
        \mathbf{\bar{C}}&=\mathbf{C}
    \end{aligned} 
\end{equation}
where $\Delta$ denotes the sampling timescale parameter and $\mathbf{\bar{A}}$ and $\mathbf{\bar{B}}$ denote the discrete counterparts of the continuous parameters $\mathbf{A}$ and $\mathbf{B}$. 
In practice, we refine the approximation of $\mathbf{\bar{B}}$ using the first-order Taylor series:
\begin{equation}
\mathbf{\bar{B}}=(exp(\mathbf{\Delta{A}})-I)(\mathbf{A})^{-1}\mathbf{B}\approx{(\mathbf{\Delta{A}})(\mathbf{\Delta{A}})^{-1}(\mathbf{\Delta{B}})}=\mathbf{\Delta{B}}
\end{equation}
Based on this approximation, the discretized state-space model (SSM) equations are reformulated to update the state and output as:
\begin{equation}
\begin{aligned}
h_{k}&=\mathbf{\bar{A}}h_{k-1}+\mathbf{\bar{B}}x_{k}\\
y_{k}&=\mathbf{\bar{C}}h_{k}+\mathbf{\bar{D}}x_{k}
\end{aligned}
\label{4}
\end{equation}
This reformulation allows the system dynamics to be captured more effectively by incorporating changes in the matrices due to small variations represented by $\mathbf{\Delta{A}}$ and $\mathbf{\Delta{B}}$.
Diverging from the time-invariance characteristic in traditional SSMs, Mamba~\cite{gu2023mamba} incorporates a selective scan mechanism (S6), ensuring that all parameters dynamically correspond to the input. To be specific, the parameters $\mathbf{B}\in{\mathbb{R}^{B\times{L}\times{N}}}$, $\mathbf{C}\in{\mathbb{R}^{B\times{L}\times{N}}}$ and $\mathbf{\Delta}\in{\mathbb{R}^{B\times{L}\times{D}}}$ are derived from the transformation of input $x\in{\mathbb{R}^{B\times{L}\times{D}}}$. This adaptation endows the model with dynamic features conducive to sequence-aware modeling. 

\paragraph{VMamba}
The VMamba~\cite{liu2024vmamba} architecture is developed on the Mamba by replacing causal 1D convolution with depth-wise 2D convolution for spatial feature extraction. It also implements the 2D selective scan (SS2D) technique, which allows parallel scanning across four directions. SS2D unfolds the image in both $HW$ and $WH$ orientations, generating four parallel sequences through flip transformations that are then processed by the SSM for selective scanning.  This approach enables each point to gather information from various directions, preserving linear complexity and significantly improving the representation of two-dimensional features.
Notably, VMamba has been thoroughly compared with previous Transformer-based backbones~\cite{dosovitskiy2020image,liu2021swin,zhang2023hivit,touvron2021training}, both theoretically and experimentally, demonstrating the effectiveness of the proposed SS2D, inspiring a significant body of subsequent works~\cite{chen2024res, shi2024multi, huang2024localmamba}.
We introduce the temporal interactions and cross-modal designs to the vanilla VMamba structure, enhancing its capabilities to tackle the complexities of audio-visual segmentation. 

\section{Methods}
\subsection{Overview}
As depicted in Fig.~\ref{fig:2}, our AVS-Mamba model is designed to generate segmentation masks $\{{S}^{t}_{m}\}^{T}_{t=1}$ for sound-emitting objects in each video frame, based on the input video sequences $\{{V}_{t}\}^{T}_{t=1}$ and their corresponding auditory cues $\{A_{t}\}^{T}_{t=1}$, where $T$ denotes the sequence length. The initial step involves the use of visual and audio encoders to extract features from the video and audio sequences, represented as $\{F^{i}_{v}\}^{4}_{i=1}$ (with $F^{i}_{v}\in{\mathbb{R}^{T\times{{h}_{i}}\times{w_{i}}\times{C}}}$, where $h_{i}$ and $w_{i}$ are the height and width of feature maps from the $i$-th scale, and $C=256$ indicates the projection dimension) and $F_{a}\in{\mathbb{R}^{T\times{1}\times{C}}}$. These multi-scale visual features $\{F^{i}_{v}\}^{4}_{i=2}$ are then processed by the Multi-scale Temporal Encoder (MTE), enhancing the capture of both intra- and inter-frame details across scales. The audio features $F_{a}$, along with the updated visual features $\{F^{i}_{v'}\}^{4}_{i=2}$, are further integrated into the Modality Aggregation Decoder (MAD), promoting the fusion of frame- and temporal-level data and enabling an implicit localization of sound-emitting objects within spatial-temporal dimensions.  Additionally, the Contextual Integration Pyramid (CIP) is utilized for cross-frame linking and upsampling of visual features $\{F^{i}_{v'}\}^{4}_{i=2}$ and $F_{v}^{1}$, along with the integration of audio features $\bar{F}_{a}\in{\mathbb{R}^{T\times{C}}}$ transformed from $F_{a}$, to refine the mask features $\mathcal{F}\in{\mathbb{R}^{T\times{h_1}\times{w_1}\times{C}}}$. The final masks $\{{S}^{t}_{m}\}^{T}_{t=1}$ are generated by the segmentation head using those enhanced mask features and the output audio queries $F_{a'}\in{\mathbb{R}^{T\times{1}\times{C}}}$.
 
\subsection{Multi-scale Temporal Encoder}
\begin{figure}
\centering
\vspace{-2mm}
\includegraphics[width=0.97\linewidth]{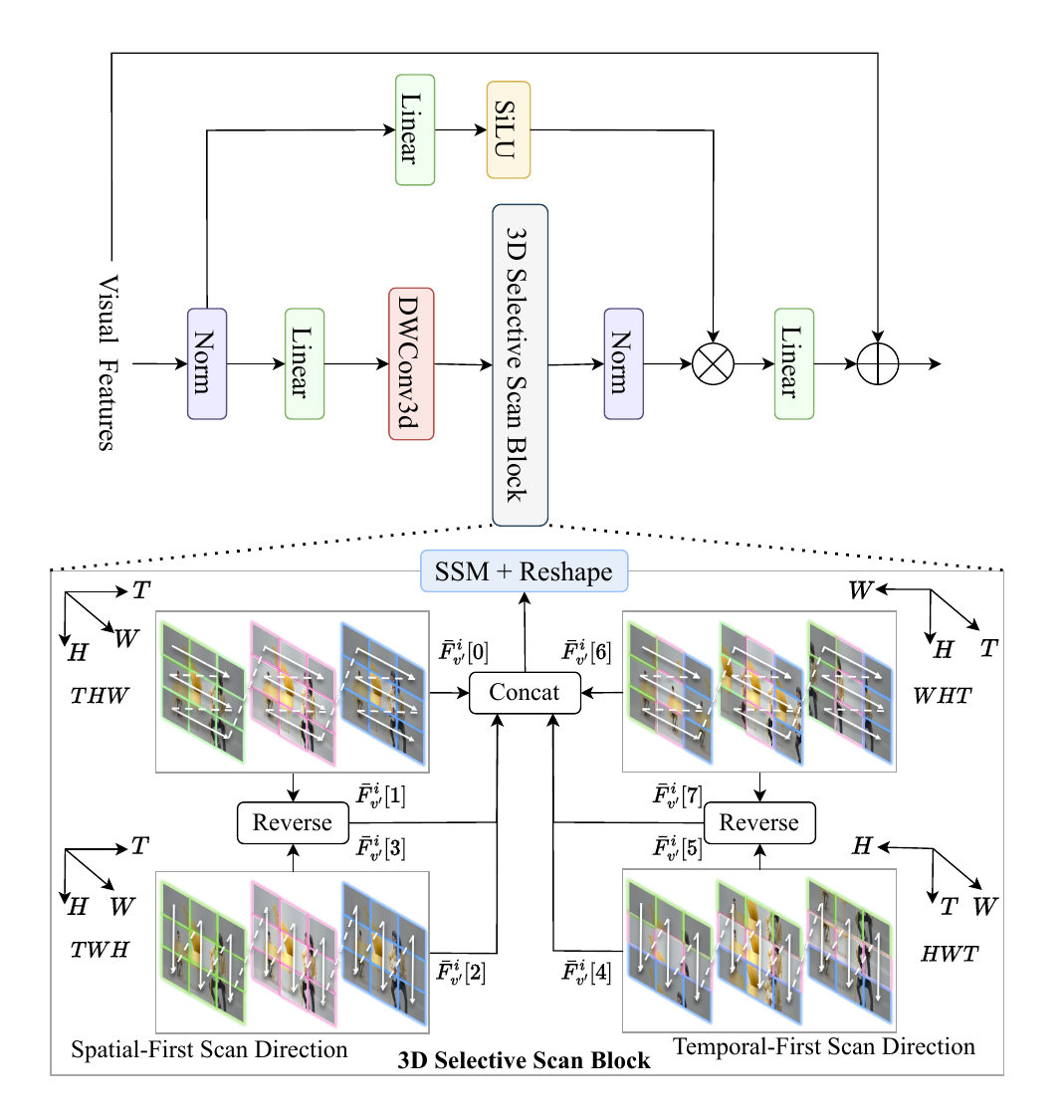}
\vspace{-3mm}
    \caption{The architecture of the Temporal Mamba Block. Initially, the video data undergoes 3D depth-wise convolution for local feature extraction, followed by modeling temporal relations using the 3D Selective Scan technique. The resulting data are multiplied by the weight parameters and passed through a residual connection, yielding visual features with enhanced temporal awareness.
    % We first employ the 3D depth-wise convolution for local feature extraction, followed by serialization 
    % Considering the video sequence $\mathcal{V}\in{\mathbb{R}^{T\times{H}\times{W}\times{C}}}$, we first employ the 3D depth-wise convolution for local feature extraction, followed by serialization of visual features through eight unfolding directions ($THW(+)(-)$, $TWH(+)(-)$, $WHT(+)(-)$ and $HWT(+)(-)$), before feeding them into the SSM for sequence modeling. This approach facilitates the central feature points collecting information from various directions, mitigating Mamba's sensitivity to orientation.
    }
    \label{fig::3}
    \vspace{-2mm}
\end{figure}
% Previous methods~\cite{gao2023avsegformer, chen2024bootstrapping} leverages the deformable transformer~\cite{zhu2020deformable} for per-frame feature collaborations across scales. However, this approach restricts interactions to merely sampled points, overlooking the extraction of spatial information from different locations within the whole sequence. Building on this, we propose the Cross-scale Mamba Encoder to establish a complementary and synergistic mechanism for intra- and inter-frame information communication through vanilla VMamba and Temporal VMamba architecture. 
Utilizing deformable transformer~\cite{zhu2020deformable} to feature scaling communication is a common approach in previous works on audio-visual segmentation, such as~\cite{gao2023avsegformer, chen2024bootstrapping}. However, this method is limited by the interactions among the sampled points. To address this limitation, we introduce the Multi-scale Temporal Encoder, which leverages the Mamba framework to enhance cross-scale collaborative interactions. Specifically, we process the projected multi-scale visual features $\{F^{i}_{v}\}^{4}_{i=2}$ through a VSS Block that uses shared weights for intra-frame modeling. Subsequently, we employ the Temporal Mamba Block to facilitate seamless integration of features across frames, the details of which are elaborated below.
 % The details are introduced as follows:

\paragraph{VSS Block}
As depicted in Fig.~\ref{fig:2}, the VSS Block is utilized to facilitate intra-frame interaction for multi-scale feature modeling. Specifically, a unified model structure with consistent weights is employed to handle features across various scales, with adjustments made primarily during the SS2D phase. In this stage, multi-scale feature maps $\{F^{i}_{v}\}^{4}_{i=2}$ ($F_{v}^{i}\in{\mathbb{R}^{T\times{h_i}\times{w_i}\times{C}}}$) are flattened and concatenated in both $HW$ (height by width) and $WH$ (width by height) orientations. This results in a concatenated feature map $F_{con}\in{\mathbb{R}^{T\times{2}\times{M}\times{C}}}$, where $M=\sum^{4}_{i=2}{h_{i}w_{i}}$ represents the combined dimensions of the feature maps.

The sequence is first inverted and then concatenated to create scanning sequences, represented as: 
\begin{equation}
    F_{SSM}=Concat(F_{con}, Rev(F_{con}))
\end{equation}
Here, $Concat$ represents the concatenation operation, $Rev$ denotes the reverse operation along the sequence dimension, and $F_{SSM}\in{\mathbb{R}^{4\times{T}\times{M}\times{C}}}$ represents the features that are input into the SSM.  

% Following this, the scanning sequences undergo disentanglement and recombination, eventually being reconstituted into the initial feature configurations $\{F^{i}_{s'}\}^{4}_{i=2}$. 
After initial preparation,  the scanning sequences $F_{SSM}$ undergoes parallelized causal feature processing as Eq.~\ref{4} through the SSM, 
the detailed process of which can be formulated as: 
\begin{equation}
    F_{s}=\sum^{4}_{k=1}{Reshape_{k}(SSM_{k}(F_{SSM}))}
\end{equation}
\begin{equation}
F^{2}_{s}, F^{3}_{s}, F^{4}_{s}=Split(F_{s})
\end{equation}
where $SSM_{k}$ refers to the application of the State Space Model to the $k$-th sequence, and  $Reshape_{k}$ is the operation that reshapes the output of $SSM_{k}$. The function $Split$ is used to divide and reshape the composite sequence $F_{s} \in \mathbb{R}^{T \times M \times C}$ back into its constituent components, aligning them with their original dimensions.

% where $Concat$ represents the concatenation operation, $Flat_{HW}$ and $Flat_{WH}$ indicate the operations that flattening along the $HW$ and $WH$ dimensions respectively and concatenating them. $Rev$ denotes reverse operation along the sequence dimension. $SSM_{k}$ and $Reshape_{k}$ refers to the State Space Model and the operation of reshape for the $k$-th sequence separately. $Split$ denotes splitting and reshaping the mixed sequence $F_{s'}\in{\mathbb{R}^{T\times{M}\times{C}}}$ back into original shape. 
% Subsequently, the output multi-scale feature maps $\{\hat{F}^{i}_{v}\}^{4}_{i=2}$ are further processed via the Temporal Mamba module, facilitating video-level interactions that result in the final output $\{F^{i}_{v'}\}^{4}_{i=2}$.

\paragraph{Temporal Mamba Block}
Previous methods~\cite{gao2023avsegformer, huang2023discovering, chen2024bootstrapping} mainly focus on cross-scale fusion within individual frames, which does not sufficiently enhance the semantic coherence of spatiotemporal features. To address these limitations, we introduce the Temporal Mamba Block to facilitate video-level multi-scale interactions. This module significantly enhances inter-frame connections among feature maps, thereby improving the overall coherence and utility of the extracted features.
% \textit{\textbf{VSS Block.}} % \label{temporal vmamba}
Fig.~\ref{fig::3} presents the architecture of the Temporal Mamba Block, which builds on the VMamba~\cite{liu2024vmamba} by incorporating depth-wise 3D convolution and the 3D Selective Scan Block to enhance the modeling of the temporal dimension. 
It processes visual features $\{\hat{F}^{i}_{v}\}^{4}_{i=2}$ received from the VSS Block, utilizing 3D depth-wise convolution to distill extract temporal-spatial characteristics effectively. 

Here we provide detailed information about the specific process of 3D Selective Scan. 
The input features are initially unfolded into 1D sequences $F_{d}\in{\mathbb{R}^{4\times{M'}\times{C}}}$, where $M'=\sum^{4}_{i=2}{h_{i}w_{i}T}$, aligning with both spatial-first ($THW$, $TWH$) and temporal-first ($HWT$, $WHT$) orders. Then $F_d$ is reorganized into eight unique scan sequences $F_{d'}\in{\mathbb{R}^{8\times{M'}\times{C}}}$, which can be represented as: 
% before being reorganized into eight unique configurations $F_{d'}\in{\mathbb{R}^{8\times{M'}\times{C}}}$. 
\begin{equation}
F_{d'}=Concat(F_{d},Rev(F_{d}))
\end{equation}
This strategy of reordering enhances the interaction of each feature point with information along both forward and backward paths. We employ the State Space Model (SSM) to parallelize the traversal of eight scanning sequences, which are subsequently restored to their original states as: 
\begin{equation}
F_{out}=\sum^{8}_{k=1}{Reshape_{k}(SSM_{k}(F_{d'}))}
\end{equation}
\begin{equation}
F^{2}_{out}, F^{3}_{out}, F^{4}_{out}=Split(F_{out})
\end{equation}
where $F_{out}\in{\mathbb{R}^{M'\times{C}}}$ represents the reshaped and merged sequence and $\{F^{i}_{out}\}^{4}_{i=2}$ represents the reorganized feature maps. 
The output feature maps $\{F^{i}_{v'}\}^{4}_{i=2}$ from the encoder will be fed into MAD and CIP for subsequent interactions. 

\begin{figure}
\centering
% \vspace{-2mm}
\includegraphics[width=0.85\linewidth]{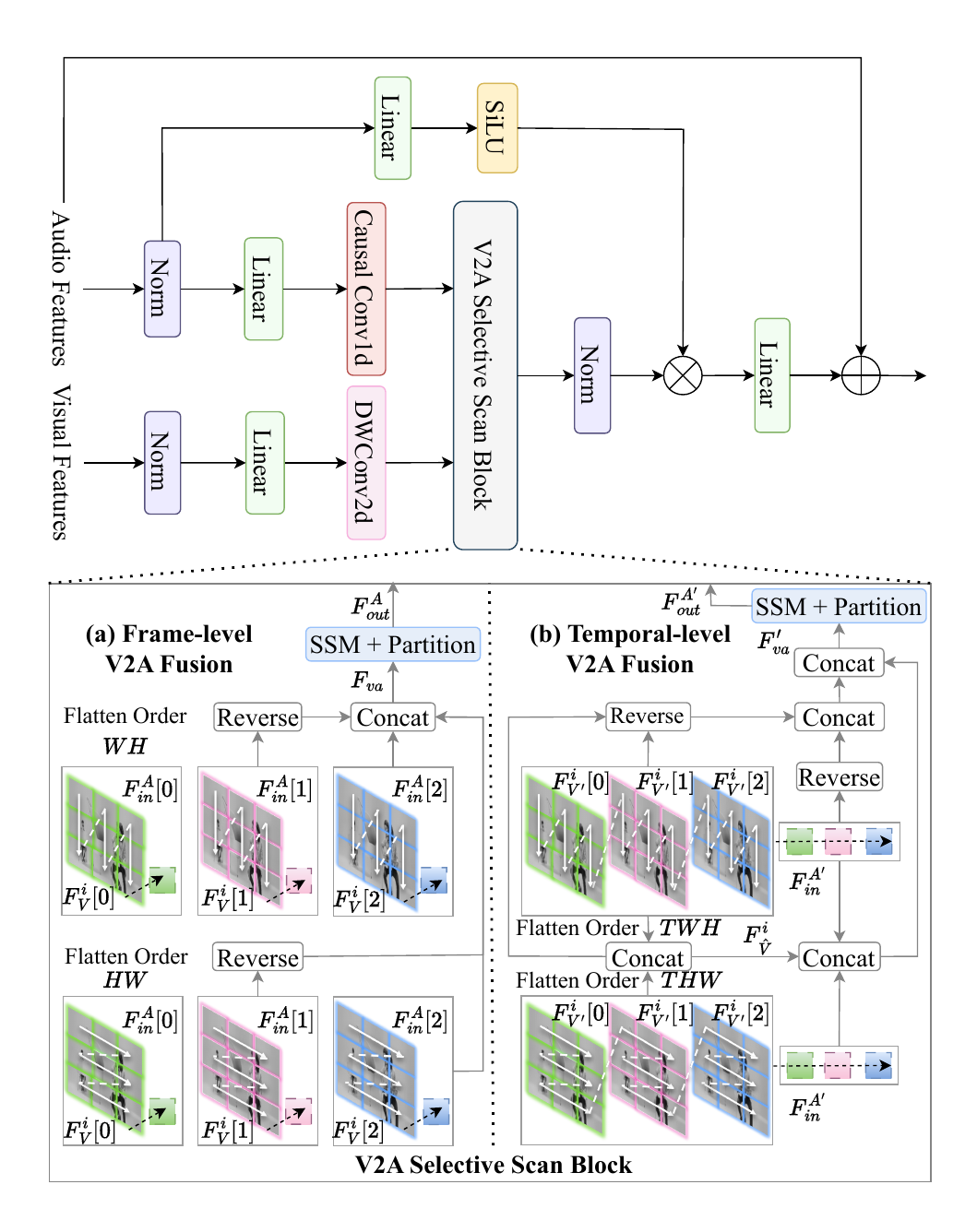}
\vspace{-3mm}
    \caption{
The architecture of the Vision-to-Audio Fusion Block. Initially, the audio feature undergoes linear mapping followed by causal 1D convolution. Subsequently, the processed audio interacts with visual features at two levels: (i) frame-level and (ii) temporal-level, through the V2A Selective Scan Block. The resulting fused audio features are then combined with the original features via weighted multiplication and seamlessly integrated into the network using a residual connection.
    }
    \label{fig::4}
\vspace{-4mm}
\end{figure}

\subsection{Modality Aggregation Decoder}
Prior research in the field~\cite{zhou2022audio, gao2023avsegformer, huang2023discovering, li2023catr} has predominantly concentrated on multi-modal interactions at the frame or temporal level, often neglecting consistency information within spatial-temporal contexts. 
To cope with these limitations, we introduce the Modality Aggregation Decoder that effectively manages frame-level and temporal-level vision-audio fusion with an audio query mechanism. This strategy aligns audio features with their corresponding sound sources and maintains continuity across frames, thereby significantly enhancing segmentation accuracy in dynamic scenes.

\paragraph{Frame-level Vision-to-Audio Fusion}
% To facilitate the aggregation of cross-modal features, we develop the Vision-to-Audio Fusion VMamba module as depicted in Fig.~\ref{fig::4} (a). 
% We first utilize the Vision-to-Audio Fusion block to identify spatial visual features within corresponding frames, as depicted in Fig.~\ref{fig::4}. 
We first utilize the Vision-to-Audio Fusion Block to perform spatial feature collaborations within corresponding frames. 
As depicted in Fig.~\ref{fig::4}, features from both the visual and audio modalities are initially subjected to linear mapping. Subsequently, a depth-wise 2D convolution is applied to visual sequences to model spatial relations, while causal 1D convolution is used for audio sequences to capture temporal relations:
% \begin{equation}
% v^{i}=DWConv2d(Linear(Norm(F^{i}_{v'})))
% \end{equation}
\begin{equation}
F^{i}_{V}=DWConv2d(Linear(Norm(F^{i}_{v'}))
\end{equation}
% \begin{equation}
% A_{in}=Conv1d(Linear(Norm(F_{a})))
% \end{equation}

\begin{equation}
F^{A}_{in}=Conv1d(Linear(Norm(F_{a})))
\end{equation}

where $Linear$ represents the linear projection, $Norm$ represents the layer normalization~\cite{ba2016layer}, $DWConv2d$ represents the depth-wise 2D convolution, $Conv1d$ represents the causal 1D convolution,  $F^{i}_{V} \in \mathbb{R}^{T \times h_{i} \times w_{i} \times C}$ and $F^{A}_{in} \in \mathbb{R}^{T \times 1 \times C}$ represent the processed visual and audio features respectively. 

Subsequently, the V2A Selective Scan method is employed to facilitate cross-modal interactions. Specifically, visual features are serialized using cross-scan strategy~\cite{liu2024vmamba}, with audio data positioned at the end of the sequence, resulting in $F_{va} \in \mathbb{R}^{4 \times T \times (h_{i}w_{i} + 1) \times C}$:
\begin{equation}
   F_{va} = Concat(Flatten(F_{V}^{i}), Repeat(F^{A}_{in}))
\end{equation}
where $Flatten$ represents the cross-scan flattening approach and $Repeat$ represents the replication of audio features. 
The four sequences in $F_{va}$ are then processed through the SSM, which effectively gathers features from vision to audio:
\begin{equation}
     F^{A}_{out} = \sum^{4}_{k=1}{Partition_{k}(SSM_{k}(F_{va}))}
\end{equation}
where $Partition_{k}$ refers to the operation that separates the audio query from the $k$-th output sequence.
Finally, the partitioned audio features $F^{A}_{out}\in{\mathbb{R}^{T\times{1}\times{C}}}$ are multiplied by the weights derived from $F_{a}$, and following linear projection and residual connection, the final output $\hat{F}_{a}\in{\mathbb{R}^{T\times{C}}}$ is produced: 
\begin{equation}
W_{a}=SiLU(Linear(Norm(F_{a})))
\end{equation}
\begin{equation}
\hat{F}_{a}=Linear(Norm(F^{A}_{out})\otimes{W_{a}})+F_{a}
\end{equation}
where $SiLU$ represents the SiLU activation function.  $W_{a}$ represents the gate weights produced by $F_{a}$. 
% Likewise, for audio-to-vision fusion, the key difference lies in positioning visual information as the main data stream. During the application of the Cross-modal Scan technique, audio data is positioned at the start of the sequence. Following the SSM processing, visual information, sorted in various orders, is consolidated and subsequently multiplied with visual weights for integration. 

\paragraph{Temporal-level Vision-to-Audio Fusion}
% After modeling the intra-frame relationships, we utilize the identical core components for video-level cross-modal interactions between $\{F^{i}_{v'}\}^{4}_{i=2}$ and $\hat{F}_{a}$. 
After establishing the intra-frame relationships, we initiate video-level cross-modal interactions between the visual feature $\{F^{i}_{v'}\}^{4}_{i=2}$ and the audio feature $\hat{F}_{a}$. 
A key distinction between frame-level fusion and temporal-level fusion is the specific modifications made to the V2A Selective Scan Block.
As shown in Fig.~\ref{fig::4}, 
following the convolution step, we unfold the visual features $F_{V'}^{i} \in {\mathbb{R}^{T \times h_{i} \times w_{i} \times C}}$ into 
$F_{\hat{V}}^{i} \in {\mathbb{R}^{2 \times Th_{i}w_{i} \times C}}$ based on the $THW$ and $TWH$ orientations. 
% \begin{equation}
% \hat{v}^{i}=Concat(flat_{THW}(v'^{i}),flat_{TWH}(v'^{i}))
% \end{equation}
% where $flat_{THW}$ and $flat_{TWH}$ denote the flattening operations along $THW$ and $TWH$ dimensions.  

We then append the audio features $F^{A'}_{in} \in \mathbb{R}^{T \times C}$ at the end of the sequence, with subsequent steps involving reversing both visual and audio features, resulting in four distinct sequences $F'_{va} \in \mathbb{R}^{4 \times (T h_{i} w_{i} + T) \times C}$. 
% Subsequent operations reverse both visual and audio features, resulting in four distinct sequencing methods $S'_{in} \in \mathbb{R}^{4 \times (T h_{i} w_{i} + T) \times C}$. % which can be formulated as:
% \begin{equation}
% S'_{in} = Concat(Concat(\hat{v}^{i}, A'_{in}), Concat(Rev(\hat{v}^{i}), Rev(A'_{in})))
% \end{equation}
During the causal traversal by the SSM, these configurations allow the audio features to establish spatial and temporal connections across visual frames, facilitating the integration of sequential information from preceding and subsequent audio frames: 
\begin{equation}
F^{A'}_{out}=\sum^{4}_{k=1}{Partition_{k}(SSM_{k}(F'_{va}))}
\end{equation}
After multiplication and linear mapping, the enhanced audio features $F^{A'}_{out} \in \mathbb{R}^{T \times C}$ produce the final output $F_{a'}$.

% This frame and video level interaction process is iterated $N$ times, subsequently utilizing the resultant audio features for crafting the ultimate mask $\{M_t\}^{T}_{t=1}$.

\begin{figure}
\centering
\vspace{-2mm}
\includegraphics[width=1\linewidth]{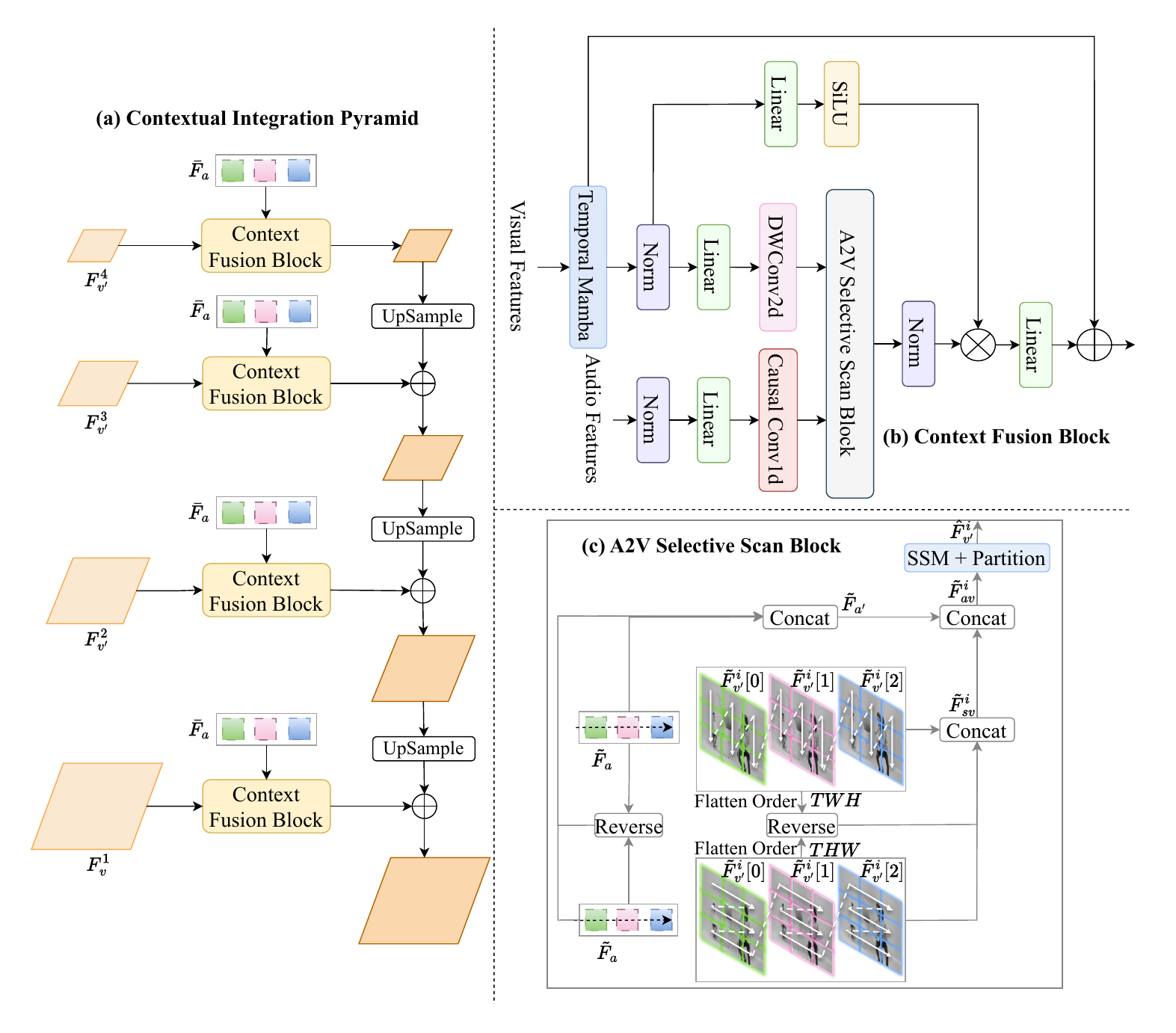}
\vspace{-4mm}
    \caption{
    The architecture of the Contextual Integration Pyramid. (a) Cross-frame audio-to-vision accumulation is performed using the Context Fusion Block, which is coupled with bilinear upsampling to align feature resolution. (b) Within each Context Fusion Block, the Temporal Mamba is employed to facilitate the exchange of temporal information across different scales of visual features. (c) We then introduce the A2V Selective Scan Block, which integrates standard audio features into the visual features through the State Space Model.
    % (a) This structure performs cross-frame audio-to-vision accumulation through the Context Fusion Block and incremental upsampling layer. (b) Within each Context Fusion Block, we adopt the Temporal Mamba to facilitate the temporal information exchange of each scale's visual features. Then we propose the (c) A2V Selective Scan Module to integrate the vanilla audio features into the visual features through SSM. 
    }
    \label{fig::cip module}
\vspace{-2mm}
\end{figure}

\subsection{Contextual Integration Pyramid}    
To further refine the multi-scale visual features and enrich them with auditory signals, we introduce the Contextual Integration Pyramid. 
As depicted in Fig.~\ref{fig::cip module} (a), it takes the initial audio features $\bar{F}_{a}$, MTE outputs $\{F^{i}_{v'}\}^{4}_{i=2}$ and $F^{1}_{v}$ as inputs. Cross-frame interactions and audio integration of each scale's visual features are facilitated via the Context Fusion Block. Subsequently, it combines multi-scale features using progressive upsampling to generate the final mask feature $\mathcal{F}$. Fig.~\ref{fig::cip module} (b) provides a detailed view of the Context Fusion Block. Within this block, input visual features first pass through the Temporal Mamba, which enhances cross-frame interactions among the visual features.  This step results in the production of feature maps $\{\tilde{F}^{i}_{v}\}^{4}_{i=1}$, which exhibit improved spatial-temporal coherence at each scale. To be specific, after the depth-wise 2D convolution, we rearrange the data into $\bar{F}^{i}_{v'}\in{\mathbb{R}^{8\times{Th_{i}w_{i}}\times{C}}}$ according to the scan direction as shown in Fig.~\ref{fig::3} and integrate it into the SSM for intricate feature modeling. Subsequently, we separate and reshape the output sequences back into $\tilde{F}^{i}_{v}\in{\mathbb{R}^{T\times{h_{i}}\times{w_{i}}\times{C}}}$, following the combination strategy of VMamba~\cite{liu2024vmamba}. This approach enables the single-scale features to benefit from an inter-frame contextual information flow in diverse orientations, thus improving the connectivity between adjacent feature points.

% This structure takes the initial audio features $\bar{F}_{a}$, MTE outputs $\{F^{i}_{v'}\}^{4}_{i=2}$ and $F^{1}_{v}$, first employing the Temporal Mamba for cross-frame interactions among visual features, thereby producing feature maps $\{\tilde{F}^{i}_{v}\}^{4}_{i=1}$ with enhanced spatial-temporal coherence within each scale.  
% To be specific, after the depth-wise 2D convolution, we rearrange the data into $\bar{F}^{i}_{v'}\in{\mathbb{R}^{8\times{Th_{i}w_{i}}\times{C}}}$ according to the scan direction as shown in Fig.\ref{fig::3}  and integrate it into the SSM for intricate feature modeling. Subsequently, we reshape the output sequences back into $\tilde{F}^{i}_{v}\in{\mathbb{R}^{T\times{h_{i}}\times{w_{i}}\times{C}}}$, following the combination strategy of VMamba~\cite{liu2024vmamba}. This approach enables the single-scale features to benefit from an inter-frame contextual information flow in diverse orientations, thus improving the connectivity between adjacent feature points.

\definecolor{lightskyblue}{rgb}{0.53, 0.81, 0.98}
\begin{table}[!tbp]
  \centering
  \caption{Comparison with state-of-the-art methods on the S4 and MS3 settings. }
  \label{tab:comparison with others in AVSBench-object}
  \scalebox{1.02}
  {
  \begin{tabular}{p{2.3cm}<{\raggedright} p{1.4cm}<{\raggedright} p{0.6cm}<{\centering} p{0.6cm}<{\centering} p{0.6cm}<{\centering} p{0.6cm}<{\centering}}
  % \begin{tabular}{p{3.4cm}<{\raggedright} p{1.9cm}{\raggedright} p{0.85cm}<{\centering} p{0.85cm}<{\centering} p{0.85cm}<{\centering} p{0.85cm}<{\certering}}
  % \begin{tabular}{p{2.3cm} p{1.4cm} p{0.6cm}<{\centering} p{0.6cm}<{\centering} p{0.6cm}<{\centering} p{0.6cm}<{\certering}}
    \toprule  %[1.2pt]
    \multirow{2}{*}{Method}& \multirow{2}{*}{Backbone} & \multicolumn{2}{c}{S4}& \multicolumn{2}{c}{MS3}\\
     &  & $M_\mathcal{J}$ & $M_\mathcal{F}$ & $M_\mathcal{J}$ & $M_\mathcal{F}$ \\
    \midrule
    \multirow{2}{*}{TPAVI~\cite{zhou2022audio}}  &  ResNet50 & 72.8 & 84.8   & 47.9 & 57.8   \\
      &  PVT-v2 & 78.7 & 87.9 & 54.0 & 64.5 \\
    \midrule
    \multirow{2}{*}{CATR~\cite{li2023catr}}  & ResNet50  & 74.8 & 86.6 & 52.8 & 65.3 \\
     & PVT-v2 & 81.4 & 89.6 & 59.0 & 70.0 \\
    \midrule
    \multirow{2}{*}{AQFormer~\cite{huang2023discovering}}  & ResNet50  & 77.0 & 86.4 & \underline{55.7} & \underline{66.9} \\
     & PVT-v2 & 81.6 & 89.4 & 61.1 & 72.1 \\
    \midrule
    \multirow{2}{*}{PIF~\cite{xu2024each}} & ResNet50 & 75.4 & 86.1 & 53.9 & 65.4\\
     & PVT-v2 & 81.4 & 90.0 & 58.9 & 70.9 \\
    \midrule
    \multirow{2}{*}{AVSBiGen~\cite{hao2023improving}} & ResNet50 & 74.1 & 85.4 & 45.0 & 56.8\\
     & PVT-v2 & 81.7 & 90.4 & 55.1 & 66.8 \\
    \midrule
    \multirow{2}{*}{DIFFAVS~\cite{mao2023contrastive}} &ResNet50& 75.8 & 86.9 & 49.8 & 58.2\\
    & PVT-v2 & 81.4 & 90.2 & 58.2 & 70.9\\
    \midrule
    \multirow{2}{*}{BAVS~\cite{liu2024bavs}} & ResNet50 & \underline{78.0} & 85.3 & 50.2 & 62.4\\
     & PVT-v2 & 82.0 & 88.6 & 58.6 & 65.5\\
    \midrule
    \multirow{2}{*}{ECMVAE~\cite{mao2023multimodal}} & ResNet50 & 76.3 & 86.5 & 48.7 & 60.7\\
     & PVT-v2 & 81.7 & 90.1 & 57.8 & 70.8\\
    \midrule
    \multirow{2}{*}{AVSegFormer~\cite{gao2023avsegformer}} & ResNet50 & 76.4 & 86.7 & 53.8 & 65.6 \\
     & PVT-v2 & 83.1 & 90.5 & 61.3 & 73.0 \\
    \midrule
    \multirow{2}{*}{AVSAC~\cite{chen2024bootstrapping}} & ResNet50 & 76.9 & \underline{87.0} & 54.0 & 65.8\\
     & PVT-v2 & \underline{84.5} & \underline{91.6} & \underline{64.2} & \underline{76.6}\\
    \midrule
    % \rowcolor{red!30}
     & ResNet50 & 76.6 & 86.2 & 54.5 & 65.6 \\
    % \rowcolor{red!30}
    \multirow{-2}{*}{SelM~\cite{li2024selm}} & PVT-v2 & 83.5 & 91.2 & 60.3 & 71.3 \\
    \midrule
    \rowcolor{gray!20}
      &  ResNet50  & \textbf{78.6}& \textbf{88.9}& \textbf{63.9}& \textbf{74.9}\\
     \rowcolor{gray!20}
   \multirow{-2}{*}{AVS-Mamba} & PVT-v2 & \textbf{85.0} & \textbf{92.6} & \textbf{68.6}& \textbf{78.8}\\
    \bottomrule  % [1.2pt]
  \end{tabular}
}
\end{table}

Then, we employ modality-specific techniques to incrementally integrate audio features into visual features across various scales. 
Specifically, the input features $\tilde{F}^{i}_{v}$ and $\bar{F}_{a}$ undergo processing through several linear and convolution layers to produce enhanced features $\tilde{F}^{i}_{v'}$ and $\tilde{F}_{a}$. 
Unlike the V2A Fusion Block, we employ the A2V Selective Scan Block for feature collaboration, as depicted in Fig.~\ref{fig::cip module} (c). In this module, the visual features $\tilde{F}^{i}_{v'}$ are first flattened in the $THW(+)(-)$ and $TWH(+)(-)$ directions, resulting in flattened visual features $\tilde{F}^{i}_{sv}\in{\mathbb{R}^{4\times{Th_{i}w_{i}}\times{C}}}$. 
% \begin{equation}
% F^{i}_{sv}=Concat(flat_{THW}(\tilde{F}^{i}_{v'}), flat_{TWH}(\tilde{F}^{i}_{v'}))
% \end{equation}
% \begin{equation}
% \tilde{F}^{i}_{sv}=Concat(F^{i}_{sv}, Rev(F^{i}_{sv}))
% \end{equation}
% where $F^{i}_{sv}$ represents the intermediate features and $\tilde{F}^{i}_{sv}\in{\mathbb{R}^{T\times{4h_{i}w_{i}}\times{C}}}$ represents the flattened visual features. 
The audio features $\tilde{F}_{a}$ are then reversed and concatenated into $\tilde{F}_{a'}\in{\mathbb{R}^{4\times{T}\times{C}}}$, followed by positioned ahead of $\tilde{F}^{i}_{sv}$ during the serialization phase. The specific scanning process is formalized as follows: 
\begin{equation}
\tilde{F}^{i}_{av}=Concat(\tilde{F}_{a'},\tilde{F}^{i}_{sv})
\end{equation}
\begin{equation}
\hat{F}^{i}_{v'}=\sum^{4}_{k=1}{Partition_{k}(SSM_{k}(\tilde{F}^{i}_{av}))}
\end{equation}
where $\tilde{F}^{i}_{av}\in{\mathbb{R}^{4\times{(T+T{h}_{i}w_{i})}\times{C}}}$ refer to the SSM input features and $\hat{F}^{i}_{v'}$ denote the final output visual features. 
Thus, each pixel can reconstruct its features informed by the accumulated audio information during the selective scanning process of SSM.
% as illustrated in Alg.~\ref{Contextual Integration Pyramid}. In detail, the input features $\tilde{F}^{i}_{v}$ and $\bar{F}_{a}$ are first processed by several linear and convolution layers. In contrast to the V2A Fusion block, the replicated audio features $\tilde{F}_{a'}$ are positioned ahead of the flattened visual features $\tilde{F}^{i}_{sv}$ during the serialization phase. Thus, each pixel can reconstruct its features informed by the accumulated audio information during the selective scanning process of SSM. Finally, the mask features $\mathcal{F}$ are generated through progressive upsampling of $\{\mathcal{F}^{i}_{v}\}^{4}_{i=1}$. 

Upon generating the mask feature $\mathcal{F}$ using the CIP module, we derive an intermediate mask $\mathcal{I}_{m}\in{\mathbb{R}^{T\times{h_{1}}\times{w_{1}}\times{1}}}$ by performing matrix multiplication between the enriched audio features $F_{a'}\in{\mathbb{R}^{T\times{1}\times{C}}}$ and $\mathcal{F}$.
Subsequently, this intermediate mask $\mathcal{I}_{m}$ is refined through a multi-layer perceptron, followed by a fully connected layer that predicts the segmentation mask for $N_{class}$ classes.

\section{Experiments}
\subsection{Datasets and Metrics}

\paragraph{Datasets}
Our study employs the AVSBench datasets for both training and evaluation, including two distinct benchmarks: AVSBench-object and AVSBench-semantic. The AVSBench-object dataset consists of two object segmentation scenarios: Single Sound Source Segmentation (S4) and Multiple Sound Sources Segmentation (MS3). Specifically,
S4 contains 4932 videos across 23 categories featuring individual sounding objects, while MS3 comprises 424 videos that feature either single or multiple sound sources. The AVSBench-semantic dataset further broadens the scope of research, encompassing 12356 videos across 70 object categories. These videos, which vary in length and include either 5 or 10 frames, are each annotated with semantic masks to classify the object categories depicted within the frames.

\paragraph{Metrics}
We utilize the Jaccard index ($M_{\mathcal{J}}$) and F-score ($M_{\mathcal{F}}$) as our primary metrics for evaluation. $M_{\mathcal{J}}$ measures the overlap by calculating the intersection-over-union (IoU) between the predicted and ground-truth segmentation masks. Conversely, $M_{\mathcal{F}}$ computes the harmonic mean of precision and recall, comprehensively assessing the model’s accuracy and sensitivity.

\begin{table}[!tbp]
 \centering
  \caption{Comparison with state-of-the-art methods on the AVSBench-semantic dataset.}
  \label{tab:comparison with others in AVSS}
  \scalebox{1.07}
  {
  \begin{tabular}{p{2.3cm}<{\raggedright} p{1.5cm}<{\centering} p{0.7cm}<{\centering} p{0.7cm}<{\centering}}
    \toprule[1.2pt]
    Method & Backbone & $M_\mathcal{J}$ & $M_\mathcal{F}$ \\
    \midrule
    \multirow{2}{*}{TPAVI~\cite{zhou2022audio}} &  ResNet50 & 20.2 & 25.2  \\
     &  PVT-v2 & 29.8 & 35.2 \\
    \midrule
    \multirow{2}{*}{BAVS~\cite{liu2024bavs}} & ResNet50 & 24.7 & 29.6\\ 
     & PVT-v2 & 32.6 & 36.4\\
    \midrule
    \multirow{2}{*}{AVSegFormer~\cite{gao2023avsegformer}} & ResNet50 & 26.6 & 31.5 \\
     & PVT-v2 & 37.3 & 42.8 \\
    \midrule
    \multirow{2}{*}{AVSAC~\cite{chen2024bootstrapping}} & ResNet50 & 25.4 & 29.7\\
    & PVT-v2 & 37.0 & 42.4\\
    \midrule
    % \rowcolor{red!30} 
    \multirow{2}{*}{SelM~\cite{li2024selm}} & ResNet50 & \underline{31.9} & \underline{37.2} \\
    % \rowcolor{red!30}
     & PVT-v2 & \textbf{41.3} & \textbf{46.9} \\
    \midrule
    \rowcolor{gray!20}
     & ResNet50  & \textbf{32.2}& \textbf{38.2}\\
    \rowcolor{gray!20}
     \multirow{-2}{*}{AVS-Mamba} & PVT-v2 & \underline{39.7}& \underline{45.1}\\
    \bottomrule[1.2pt]
    \end{tabular}
    }
  % \vspace{-3mm}
  % The best and second-best results are marked by \textbf{bold} and \underline{underline}, respectively.}
  \vspace{-3mm}
\end{table}

\subsection{Implementation Details}
For fair comparisons with prior studies~\cite{zhou2022audio,huang2023discovering,li2023catr,gao2023avsegformer}, we employ ResNet50~\cite{he2016deep}, pretrained on the MSCOCO~\cite{lin2014microsoft}, and PVT-v2~\cite{wang2022pvt}, pretrained on the ImageNet~\cite{russakovsky2015imagenet} as our visual encoder. VGGish~\cite{arandjelovic2017look} pretrained on AudioSet~\cite{gemmeke2017audio} is utilized as our audio encoder. 
% Both the Encoder and Decoder are configured with two layers each. 
The number of layers for both the encoder and decoder is set to two. 
In modules incorporating Mamba, we adhere to the parameter settings specified by VMamba~\cite{liu2024vmamba}.
% The causal 1D convolution's parameters mirror those of the original Mamba~\cite{gu2023mamba}, with the 3D convolution kernel specified as $3\times{3}\times{1}$ and a uniform drop path rate of 0.1.
% We set the number of layers in the Encoder and Decoder as two. For all modules containing Mamba, we adopt the parameter settings of VMamba. The parameter configuration for the causal 1D convolution is the same as that of the original Mamba, with the kernel size of the 3D convolution set to $3\times{3}\times{1}$, and drop path rates uniformly set to 0.1.
We employ the AdamW optimizer, setting the learning rate to 2e-5. For the AVSBench-object dataset, we utilize dice loss for supervision, whereas binary cross-entropy loss is used for the AVSBench-semantic dataset. The batch sizes during training are configured to two for AVSBench-object and six for AVSBench-semantic. Training schedules vary across datasets: 60 epochs for the S4 setting, 120 epochs for MS3, and 50 epochs for the AVSBench-semantic dataset.
Considering the number of frames varies across the AVSBench-semantic dataset, we standardize the data by padding all videos to a uniform length of 10 frames, simplifying processing and analysis.
Data augmentation strategies implemented across all datasets include horizontal flipping, random resizing and cropping, and photometric distortion, enhancing the robustness and generalizability of our model.
\begin{figure*}
    \centering
    % \vspace{-3mm}
    \includegraphics[width=0.95\linewidth]{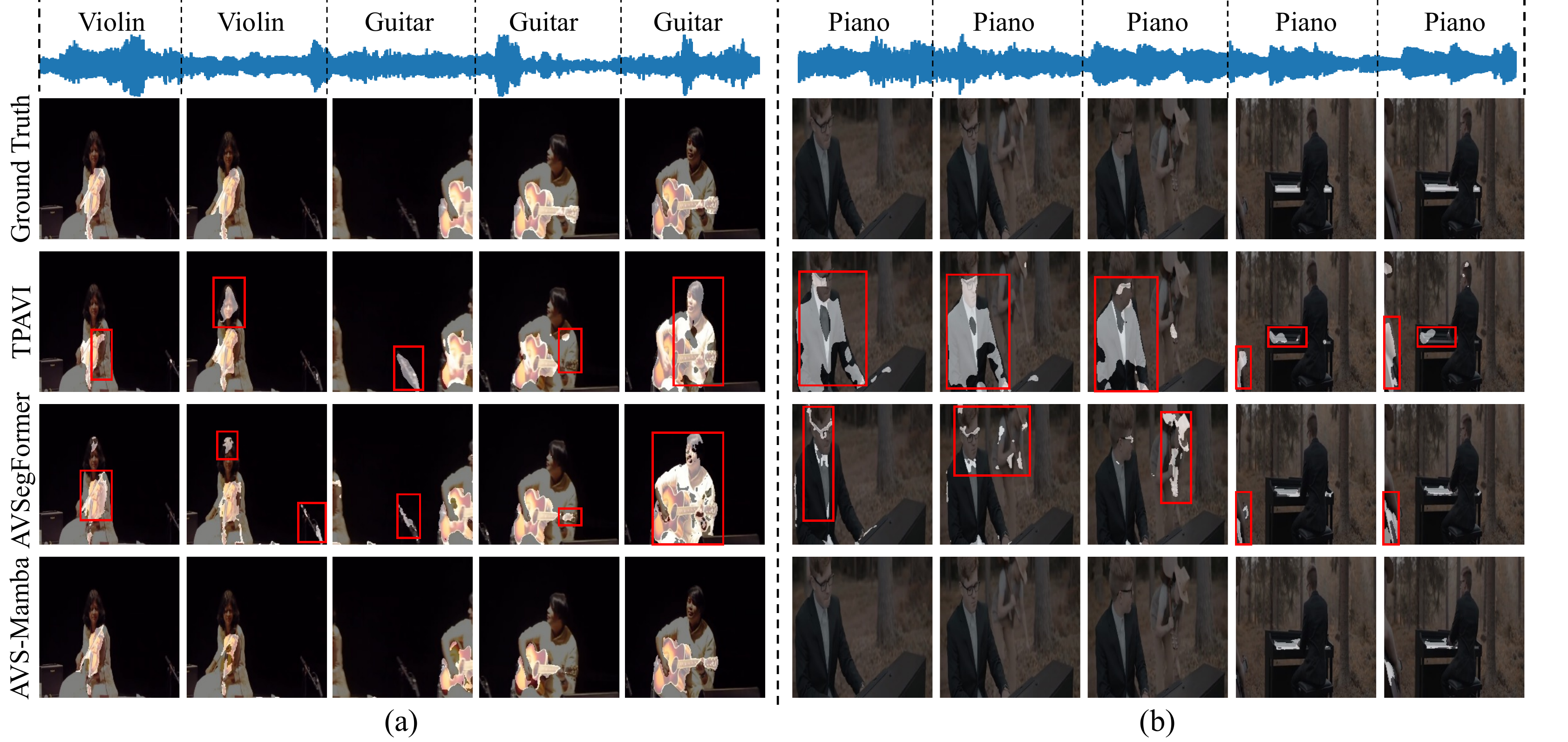}
    \vspace{-3mm}
    \caption{Qualitative comparison between TPAVI~\cite{zhou2022audio}, AVSegFormer~\cite{gao2023avsegformer} and AVS-Mamba on the AVSBench-object dataset. 
    % (a) Scenarios involving transitions of the sound sources. (b) Instances of missing sound sources. 
    Previous methods struggle to accurately capture the transitions of sound sources in dynamic scenarios (a) and fail in adapting to frames where sounding objects are absent (b). In contrast, AVS-Mamba demonstrates robust segmentation performance in these complex scenes, effectively handling the challenges presented.
    }
    \label{fig:result visualization}
    % \vspace{-2mm}
\end{figure*}

\subsection{Main Results}
Table~\ref{tab:comparison with others in AVSBench-object} presents the comparative performance of our AVS-Mamba method against previous landmark studies~\cite{li2023catr, huang2023discovering, xu2024each, hao2023improving, mao2023contrastive, liu2024bavs, mao2023multimodal, chen2024bootstrapping, zhou2022audio, gao2023avsegformer, li2024selm} on the AVSBench-object datasets. In the S4 configuration, AVS-Mamba achieves a notable improvement, outperforming existing state-of-the-art (SOTA) methods by 0.6 $M_{\mathcal{J}}$ and 1.9 $M_{\mathcal{F}}$ with a ResNet50 visual encoder, and by 0.5 $M_{\mathcal{J}}$ and 1.0 $M_{\mathcal{F}}$ when utilizing PVT-v2. In the more challenging MS3 setting, our method demonstrates significant performance improvements. Using ResNet50, we achieve gains of 8.2 $M_{\mathcal{J}}$ and 8.0 $M_{\mathcal{F}}$, while with PVT-v2, we observe gains of 4.4 $M_{\mathcal{J}}$ and 2.2 $M_{\mathcal{F}}$. These results significantly surpass those of earlier approaches.

Additionally, Table~\ref{tab:comparison with others in AVSS} offers a detailed comparison of our method with others on the AVSBench-semantic dataset. 
Notably, when utilizing both ResNet50 and PVTv2 as visual encoders, AVS-Mamba outperforms SelM~\cite{li2024selm}, another recent implementation of a Mamba-based architecture.
% Notably, when using ResNet50 as the visual encoder, AVS-Mamba achieves substantial enhancements of 5.6 $M_{\mathcal{J}}$ and 6.7 $M_{\mathcal{F}}$.
% With the PVT-v2 encoder, the improvements are 2.4 in $M_{\mathcal{J}}$ and 2.3 in $M_{\mathcal{F}}$. 
The competitive performance across both settings affirms the robustness of our task-specific enhancements to the Mamba structure and underscores the effectiveness of AVS-Mamba in facilitating multi-modal fusion.

\begin{figure*}
    \centering
    % \vspace{-2mm}
    \includegraphics[width=0.96\linewidth]{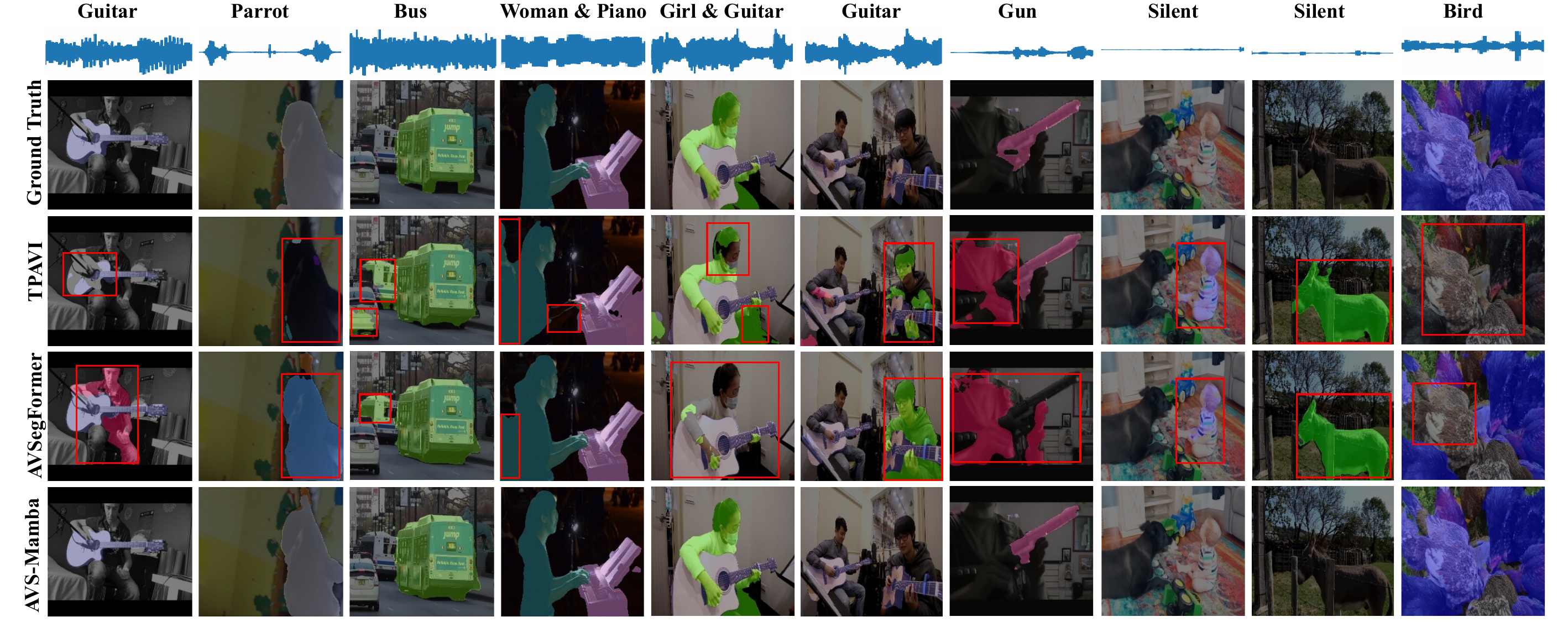}
    \vspace{-3mm}
    % 暂时没写完
    \caption{Qualitative comparison between TPAVI~\cite{zhou2022audio}, AVSegFormer~\cite{gao2023avsegformer} and AVS-Mamba on the AVSBench-semantic dataset. In comparison with earlier approaches, AVS-Mamba exhibits superior capabilities of interpreting semantic information, highlighting its more effective integration and guidance of multi-modal contextual relationships. }
    \label{fig:semantic visualization}
    \vspace{-4mm}
\end{figure*}

\begin{figure}
    \centering
 \vspace{-2mm}
    \includegraphics[width=1\linewidth]{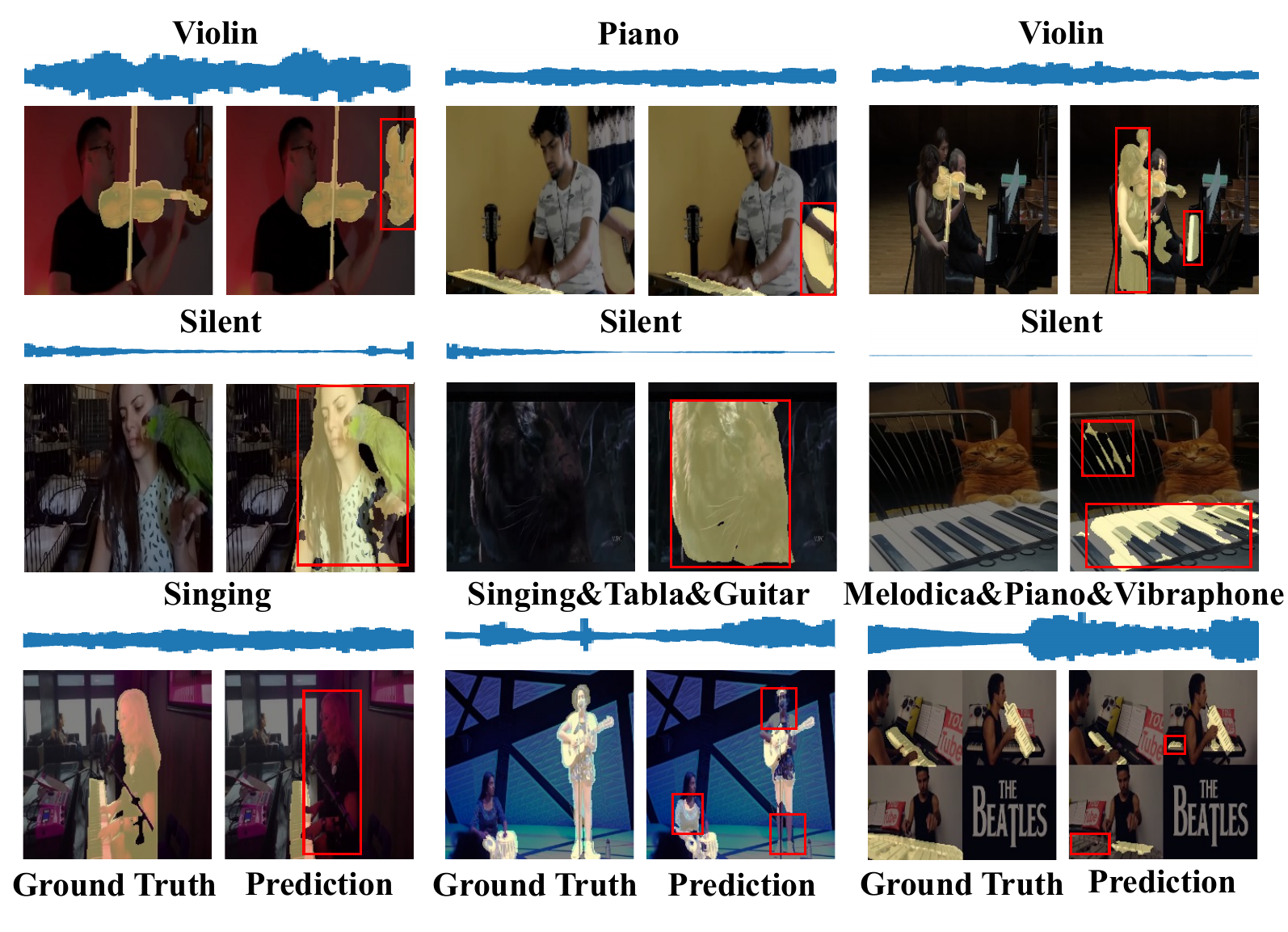}
    \vspace{-6mm}
    % 暂时没写完
    \caption{Failure case analysis for AVS-Mamba. We present and examine specific scenarios where the model underperforms, including scenes with a single sound source (top row), silent frames (middle row), and multiple-sounding objects (bottom row).}
    \label{fig:failure cases}
    \vspace{-2mm}
\end{figure}

\subsection{Qualitative Analysis}
\paragraph{Comparison of AVSBench-object}
Fig.~\ref{fig:result visualization} presents a visual comparison of AVS-Mamba with TPAVI~\cite{zhou2022audio} and AVSegFormer~\cite{ gao2023avsegformer} using PVT-v2 as the backbone. In Fig.~\ref{fig:result visualization} (a), which features complex transitions between sounding objects (from violin to guitar), both TPAVI and AVSegFormer struggle to accurately pinpoint the change in sounding objects. In contrast, AVS-Mamba utilizes progressive intra- and inter-frame feature interactions, leveraging the Mamba framework to enhance perception across adjacent frames. This allows AVS-Mamba to accurately discern changes across frames and identify the sounding objects. Furthermore, Fig.~\ref{fig:result visualization} (b) demonstrates the inadequacies of previous methods in handling frames lacking sounding objects. Unlike these methods, AVS-Mamba effectively integrates 1D auditory signals and 2D sound source regions through comprehensive cross-modal fusion, showcasing the robustness and efficacy of our approach in adapting to varied audio-visual scenarios.

% Fig.~\ref{fig:failure cases} visualize a failure instance, which showcases complex scenarios of sound source transitions, with AVS-Mamba unable to precisely segment the sound source (the lion or dinosaur) throughout the initial four frames. Moreover, in the concluding silent frame, it erroneously identifies an object as producing sound. The issue could arise from the state compression characteristic of Mamba, potentially limiting the depth of inter-frame interactions and consequently impairing accurate detection.
% indicating that our model's sensitivity towards auditory information needs further enhancement. 
% Fig.6 displays the failure instances, highlighting that our model's performance in identifying and segmenting multiple small sound-emitting objects requires further enhancement.

\paragraph{Comparison of AVSBench-semantic}
Fig.~\ref{fig:semantic visualization} provides a visual comparison of the AVSBench-semantic dataset, where different colors represent various semantic categories. In the examples shown in the $1^{st}$ and $2^{nd}$ columns, TPAVI fails to fully segment the guitar and does not recognize the sounding parrot in the second instance. AVSegFormer also incorrectly categorizes the subjects in both cases. In contrast, AVS-Mamba accurately produces complete and detailed segmentation masks, demonstrating its superior ability to integrate audio-visual features for semantic interpretation.
In the $3^{rd}$ to $5^{th}$ examples, previous methods either over-segment parts of objects or fail to segment them entirely. Additionally, results in the following two columns reveal that these methods often mistakenly identify non-vocalizing objects as emitting sounds. Conversely, AVS-Mamba precisely pinpoints the location and boundaries of sound sources.
The $8^{th}$ and $9^{th}$ columns highlight AVS-Mamba’s robustness in handling silent frames, illustrating the effectiveness of our carefully designed modules in audio signal integration. The final column presents a complex scenario involving multiple vocalizing birds. Here, TPAVI misidentifies the scene as silent, whereas AVS-Mamba accurately captures minimal omissions of vocalizing objects, outperforming AVSegFormer.  This comparison highlights our method's robust capability in extracting semantic signals from audio for effective audio-visual semantic segmentation.

\begin{figure}
    \centering
    \vspace{-2mm}
    \includegraphics[width=1\linewidth]{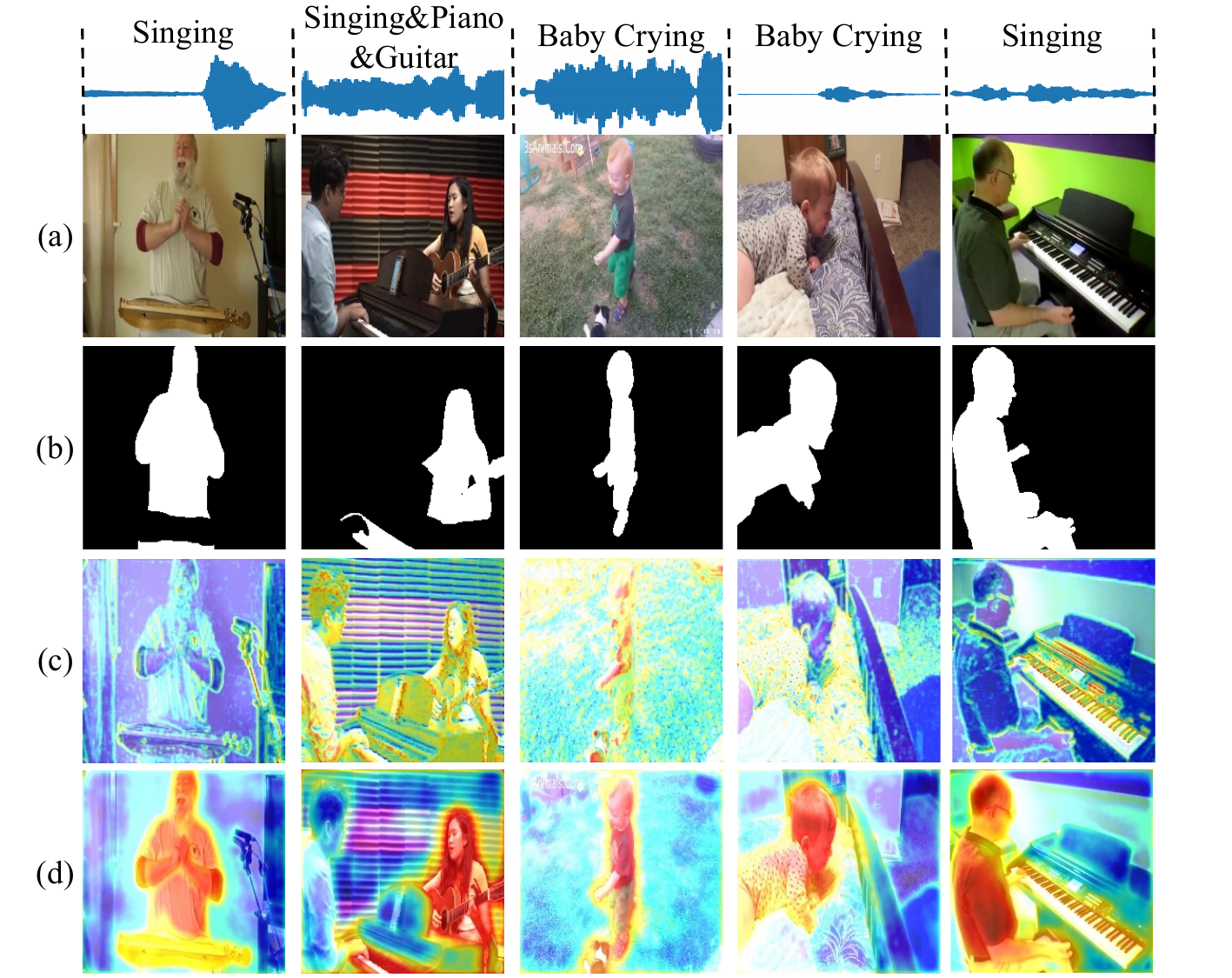}
    \vspace{-5mm}
    % 暂时没写完
    \caption{Feature map visualizations presented in sequence from top to bottom: (a) raw images, (b) ground truth masks, (c) feature maps before processing by the Contextual Integration Pyramid, and (d) feature maps after processing by the Contextual Integration Pyramid.}
    \vspace{-2mm}
    \label{fig:feature visualization}
    % \vspace{-1mm}
\end{figure}

% \subsection{Failure Case Analysis}
\subsection{Visualization Analysis}
\paragraph{Failure Case Analysis}
In Fig.~\ref{fig:failure cases}, we present several typical failure cases that highlight the current limitations of AVS-Mamba across different scenarios. From top to bottom, the figure displays challenges with a single sound source, silent frames, and multiple sound source environments. Through visual analysis, it is evident that AVS-Mamba still struggles to achieve accurate segmentation in these contexts.

In the first example, despite AVS-Mamba's use of bidirectional cross-modal fusion, visual features occasionally overshadow audio information. This imbalance can cause the model to incorrectly segment multiple instances in scenarios with a single sound source. In the second example, the model erroneously detects silent frames as containing sound-emitting objects. This error likely stems from Mamba's sequential modeling, which incorporate audio information from adjacent sounding frames into a silent frame through temporal cross-modal interactions. In the third case, AVS-Mamba struggles to accurately process environments with multiple sound sources due to the causal modeling process where certain visual features inadequately capture audio signals, leading to inactive sounding regions.
% As indicated in the $1^{st}$ row, despite AVS-Mamba’s implementation of bidirectional cross-modal fusion, visual features can sometimes dominate the integration process, leading to the oversight of audio information. Consequently, the model may incorrectly segment multiple instances in single sound source scenarios.
% In the $2^{nd}$ row, the model mistakenly identifies silent frames as containing sound-emitting objects. This might be attributed to the sequential modeling characteristics of Mamba, which can introduce redundant audio information from adjacent sounding frames into the current silent frame during temporal cross-modal interactions.
% Lastly, the visual results in the $3^{rd}$ row show that AVS-Mamba faces challenges in accurately processing environments with multiple sounding objects. 
% This occurs because, during the causal modeling process, certain visual features fail to adequately perceive audio signals, resulting in the non-activation of sounding regions.

To mitigate these issues, future work will refine AVS-Mamba’s selective scanning strategy to decrease sensitivity to feature point distances. We will also explore integrating binary matching strategies or semantic decoupling of sound sources to improve the model's capability to discern and segment multiple sound sources effectively.
% To address the above issues, we plan to optimize AVS-Mamba’s selective scanning strategy in future work to reduce the distance sensitivity between feature points. Additionally, we will implement binary matching~\cite{carion2020end} or semantic decoupling of sound sources~\cite{li2024qdformer} strategies to enhance the model’s ability to recognize multiple sound sources.}
% During Mamba’s causal modeling process, the spatial distance between visual and audio features prevents some visual features from effectively capturing audio information. Consequently, certain sounding regions are not accurately identified.}

% This misclassification occurs because, during temporal cross-modal interactions, redundant audio signals from active frames activate the image regions of sounding instances within silent frames through Mamba’s sequential scanning mechanism.

% In the $1^{st}$ row, within frames featuring a single sound source, AVS-Mamba incorrectly identifies multiple sounding objects. This might be attributed to the causal modeling characteristics of Mamba, which can introduce redundant information from adjacent frames into the current frame. In the $2^{nd}$ row, the model mistakenly identifies silent frames as containing sound-emitting objects, indicating that AVS-Mamba does not adequately utilize the audio signals within these frames. Lastly, the visual results in the $3^{rd}$ row show that AVS-Mamba faces challenges in accurately processing environments with multiple sounding objects.

\paragraph{Feature Visualization}
To intuitively demonstrate the impact of the Mamba modules, we visualize the feature maps from the visual backbone both before and after integration using the Contextual Integration Pyramid (CIP) module. As shown in Fig.~\ref{fig:feature visualization} (c) and (d), we apply the principal component analysis (PCA) technique ~\cite{shlens2014tutorial} to distill the principal channels from the feature maps. These channels are then superimposed on the original images using a scaling factor of 0.7, with regions of model focus highlighted.
This visualization reveals that, in the absence of integrated audio features, the model tends to focus primarily on background regions. However, once audio-to-vision feature aggregation is performed through the CIP module, there is a noticeable shift in the model’s focus towards areas containing sound-emitting objects. This qualitative result underscores the CIP module’s effectiveness in enhancing the Mamba’s cross-modal capabilities, demonstrating its ability to direct attention to relevant audio-visual correlations effectively.

\begin{table}[tbp]
    \centering
    \vspace{-2mm}
    \caption{Overall components analysis.
    % The outcomes demonstrate the effectiveness of the proposed modules.
    }
    \label{Ablation study of overall components}
    \resizebox{64mm}{!}{\begin{tabular}{ccccc}
    \toprule
    % 这里的各个实验及模块的名称待定
         \multirow{2}{*}{Module} &\multicolumn{2}{c}{MS3}&\multicolumn{2}{c}{S4}\\
         & \(M_{\mathcal{J}}\)&\(M_{\mathcal{F}}\) & \(M_{\mathcal{J}}\)&\(M_{\mathcal{F}}\)\\
         \midrule
         $\it{w/o}$ MTE&  66.8 & \underline{77.3}&\underline{84.3}& \underline{92.1}\\
         $\it{w/o}$ MAD& 65.4&75.7 &84.0& \underline{92.1}\\
         $\it{w/o}$ CIP& \underline{67.1}&77.0& 83.9& 91.7\\
         Ours& \textbf{68.6}& \textbf{78.8}&\textbf{85.0}&\textbf{92.6}\\
    \bottomrule
    \end{tabular}}
    \vspace{-3mm}
\end{table}
\begin{table}[tbp]
\centering
    % \vspace{-2mm}
    \caption{Analysis of Multi-scale Temporal Encoder.
    % Both the VSS block and Temporal Mamba block are pivotal in multi-scale interactions.
    }
    \label{Ablation study of Multi-scale Temporal Encoder}
   \resizebox{75mm}{!}{\begin{tabular}{ccccc}
    \toprule
    % 这里的各个实验及模块的名称待定
         \multirow{2}{*}{Components} &\multicolumn{2}{c}{MS3}&\multicolumn{2}{c}{S4}\\
         & \(M_{\mathcal{J}}\)&\(M_{\mathcal{F}}\)& \(M_{\mathcal{J}}\)&\(M_{\mathcal{F}}\)\\
         \midrule
         $\it{w/o}$ Encoder&  66.8& 77.3&84.3& 92.1\\
         $\it{w/o}$ VSS Block&  \underline{68.1}& \underline{77.4}&84.4&92.1\\
         $\it{w/o}$ Temporal Mamba& 67.6&77.2& \underline{84.6}& \underline{92.2}\\
         Ours& \textbf{68.6}&\textbf{78.8}& \textbf{85.0}& \textbf{92.6}\\
    \bottomrule
    \end{tabular}}
    \vspace{-2mm}
\end{table}
\begin{table}[!tbp]
\centering
    \vspace{-2mm}
    \caption{Analysis of Modality Aggregation Decoder.
    % Implementation of both frame-level fusion and temporal-level cross-modal fusion significantly enhances performance compared to methods that omit either of them.
    }
    \label{Ablation study of Modality Aggregation Decoder}
    \resizebox{75mm}{!}{\begin{tabular}{ccccc}
    \toprule
    % 这里的各个实验及模块的名称待定
         \multirow{2}{*}{Components}& \multicolumn{2}{c}{MS3} & \multicolumn{2}{c}{S4}\\
         &\(M_{\mathcal{J}}\)&\(M_{\mathcal{F}}\)&\(M_{\mathcal{J}}\)&\(M_{\mathcal{F}}\)\\
         \midrule
         $\it{w/o}$ Decoder &  65.4& 75.7&  84.0 & 92.1\\
         $\it{w/o}$ Frame-level &  \underline{67.0}& \underline{77.1}& 84.5& \underline{92.3} \\
         $\it{w/o}$ Temporal-level & 66.6&76.8&\underline{84.6}&92.2\\
         Ours & \textbf{68.6}&\textbf{78.8}&\textbf{85.0}&\textbf{92.6}\\
    \bottomrule
    \end{tabular}}
    \vspace{-3mm}
\end{table}

\begin{table}[tbp]
\centering
    % \vspace{-2mm}
    \caption{Ablation study of Contextual Integration Pyramid.
    % Omitting either the Temporal Mamba or the A2V Selective Scan Block results in diminished performance.
    }
    \label{Ablation study of Contextual Integration Pyramid}
    \resizebox{75mm}{!}{\begin{tabular}{ccccc}
    \toprule
          \multirow{2}{*}{Components}& \multicolumn{2}{c}{MS3} & \multicolumn{2}{c}{S4}\\
          & \(M_{\mathcal{J}}\)&\(M_{\mathcal{F}}\)& \(M_{\mathcal{J}}\)&\(M_{\mathcal{F}}\)\\
         \midrule
         % 这里的各个实验及模块的名称待定
         $\it{w/o}$ CIP&  67.1& 77.0& 83.9 & 91.7\\
         $\it{w/o}$ Temporal Mamba&  67.9& 76.8&84.6 &92.1\\
         $\it{w/o}$ A2V Selective Scan & \underline{68.5}&\underline{78.1}& \textbf{85.0} & \underline{92.4}\\
         Ours & \textbf{68.6}&\textbf{78.8}& \textbf{85.0} & \textbf{92.6}\\
    \bottomrule
    \end{tabular}}
\end{table}
\begin{table}[!tbp]
\centering
    \vspace{-2mm}
    \caption{Impact of scan order in Temporal Mamba.
    % Utilizing eight scan sequences achieves optimal results.
    }
    \label{Ablation study of scan order}
    \resizebox{60mm}{!}{
    \begin{tabular}{ccccc}
    \toprule
    % 这里的各个实验及模块的名称待定
        \multirow{2}{*}{Num.}& \multicolumn{2}{c}{MS3} & \multicolumn{2}{c}{S4}\\
         & \(M_{\mathcal{J}}\)&\(M_{\mathcal{F}}\)& \(M_{\mathcal{J}}\)&\(M_{\mathcal{F}}\)\\
         \midrule
         2&  67.6& 77.3 &84.1&91.8\\
         4&  67.8& 77.0& 84.2 &92.2\\
         6& \underline{68.7}& 77.3 & \underline{84.6}& \underline{92.3}\\
         8 (Ours)& 68.6&\textbf{78.8}&\textbf{85.0}&\textbf{92.6}\\
         10 & \textbf{69.0}& \underline{78.3}& \underline{84.6} & 92.2 \\
         12& 68.6& 78.1 & 84.3 & 92.0 \\
    \bottomrule
    \end{tabular}}
    \vspace{-3mm}
\end{table}

\begin{figure*}
    \centering
    \vspace{-5mm}
    \includegraphics[width=1\linewidth]{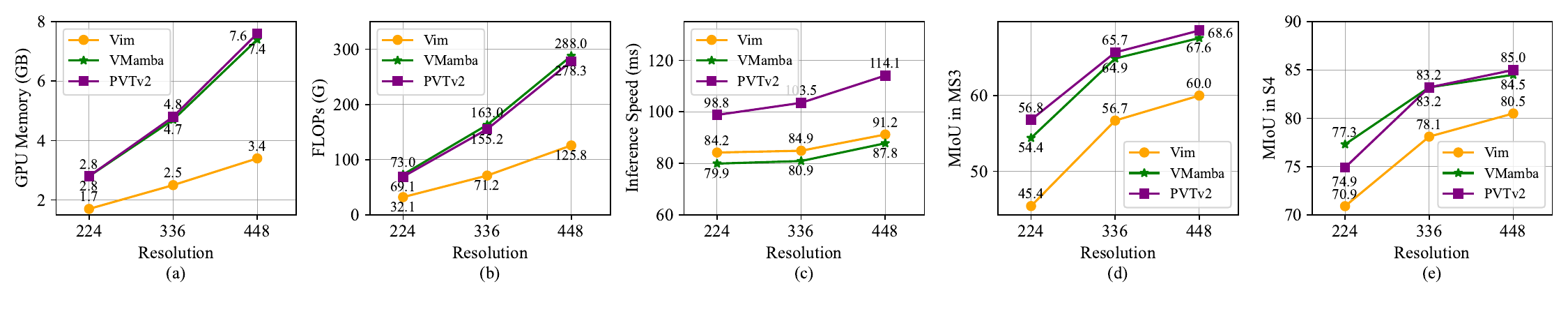}
    \vspace{-11mm}
    \caption{Comparative analysis of (a) GPU memory consumption, (b) FLOPs and (c) inference speed using Vim, VMamba, and PVTv2 as backbones. Subfigures (d) and (e) present performance metrics $M_{\mathcal{J}}$ on the AVSBench-object dataset.}
    \label{fig:plot}
    \vspace{-3mm}
\end{figure*}

\subsection{Ablation Study}
To verify the efficacy of the proposed modules, which include the Multi-scale Temporal Encoder, Modality Aggregation Decoder, and Contextual Integration Pyramid, we conduct ablative experiments using PVT-v2 as our visual encoder on the AVSBench-objects dataset. Additionally, we explore the impact of scanning sequence variations within the Temporal Mamba Block to understand their effects on the overall system performance. Finally, we conduct experiments to compare the GPU memory usage, GFLOPs, and inference speeds of AVS-Mamba using Vim and VMamba as visual encoders versus PVT-v2. These
experiments are performed across different input resolutions on the
AVSBench-object dataset.

\paragraph{Overall Components Analysis} 
We assess the impact of critical components on our model's effectiveness, specifically focusing on the Multi-modal Temporal Encoder (MTE), Modality Aggregation Decoder (MAD), and Contextual Integration Pyramid (CIP). As shown in Table~\ref{Ablation study of overall components}, the omission of any single module leads to a decline in performance. For example, removing the MAD module results in a decrease of 3.2 $M_{\mathcal{J}}$ in the MS3 setting and 1.0 $M_{\mathcal{J}}$ in the S4 setting, underscoring the significant role of the cross-modal design of the V2A Fusion Block. Similarly, eliminating the CIP module leads to a reduction of 1.5 $M_{\mathcal{J}}$ in MS3 and 1.1 $M_{\mathcal{J}}$ in S4. Furthermore, the removal of the MTE module results in a drop of 1.8 $M_{\mathcal{J}}$ in MS3 and 0.7 $M_{\mathcal{J}}$ in S4. This comprehensive ablation study verifies the critical importance and effectiveness of each component in our model's architecture.

\paragraph{Multi-scale Temporal Encoder} 
Our initial investigation focuses on the MTE module's contribution to model efficacy. As evidenced by Table~\ref{Ablation study of Multi-scale Temporal Encoder}, the elimination of either VSS Block or Temporal Mamba Block triggers a notable decrease in performance metrics. Furthermore, the exclusion of Temporal Mamba significantly impacts outcomes, with a reduction of 1.0 $M_{\mathcal{J}}$ in MS3 and 0.4 $M_{\mathcal{J}}$ in S4 observed. This highlights the critical role of Temporal Mamba in conducting temporal modeling of multi-scale features, effectively facilitating the extraction of cross-scale spatial correlations and enhancing the perception between adjacent frames.
% effectively fostering the extraction of both cross-scale spatial correlations and temporal dependencies across frames, thereby elevating segmentation accuracy.

\paragraph{Modality Aggregation Decoder} 
Table~\ref{Ablation study of Modality Aggregation Decoder} offers insights into the ablation study of the Modality Aggregation Decoder (MAD), with the outcomes underscoring the critical role of utilizing the V2A Fusion Block for cross-modal interactions at the frame and temporal level. The necessity of this approach stems from frame-level interactions enabling audio features to concentrate on diverse feature points within frames, while video-level interactions amplify the comprehension of visual information across the temporal dimension. These continuous interactions substantially advance the model’s capability in accurately decoding and perceiving sound sources.

\paragraph{Contextual Integration Pyramid} 
For the ablation study of the Contextual Integration Pyramid (CIP), as detailed in Table~\ref{Ablation study of Contextual Integration Pyramid}, we maintain the FPN framework while specifically investigating the effects of removing the Temporal Mamba Block and A2V  Selective Scan Block. Our observations indicate that omitting either the Temporal Mamba Block or the A2V Selective Scan Block leads to a decline in performance. Notably, removing the Temporal Mamba Block, which is critical for inter-frame interactions, results in a significant performance drop of 0.7 $M_{\mathcal{J}}$ in MS3 and 0.4 $M_{\mathcal{J}}$ in S4. This underscores the importance of fine-grained fusion across frames.

\paragraph{Number of Scan Orders}
We also examine the effect of varying the number of scanning sequences in the Temporal Mamba Block. We test configurations of 2 ($THW(+)(-)$), 4 ($THW(+)(-)$, $TWH(+)(-)$), 6 ($THW(+)(-)$, $TWH(+)(-)$, $HWT(+)(-)$), 8 ($THW(+)(-)$, $TWH(+)(-)$, $HWT(+)(-)$, $WHT(+)(-)$), 10 ($THW(+)(-)$, $TWH(+)(-)$, $HWT(+)(-)$, $WHT(+)(-)$, $HTW(+)(-)$) and 12 ($THW(+)(-)$, $TWH(+)(-)$, $HWT(+)(-)$, $WHT(+)(-)$, $HTW(+)(-)$, $WTH(+)(-)$) directions. The results, presented in Table~\ref{Ablation study of scan order}, indicate that using 8 scanning directions optimizes performance, likely due to the Mamba framework's directional and sequential sensitivity. This setup allows for more effective integration of temporal-spatial information at each point.

Increasing scanning sequences beyond 8 offers marginal benefits and, in some cases, diminishes performance. For instance, expanding to 10 directions slightly improves the $M_\mathcal{J}$ metric on the MS3 dataset but reduces performance on other metrics. Extending to 12 directions consistently reduces performance across all metrics, suggesting that extra directions do not linearly improve segmentation accuracy. This indicates that the additional directions do not linearly enhance segmentation accuracy and that the initial eight orientations, covering both temporal-first and spatial-first scenarios, are optimal. Additional orientations, like $HTW$ and $WTH$, disrupt the model's causal modeling by excessively blending row and column features across frames.

\paragraph{Exploration of the Variants of AVS-Mamba}
As illustrated in Fig.~\ref{fig:plot} (a-c), employing Vim as the backbone results in the lowest GFLOPs and GPU memory consumption, and significantly faster inference speeds compared to using PVT-v2. However, the performance metrics shown in Fig.~\ref{fig:plot} (d-e) indicate that Vim's approach of downsampling images to 1/16 of their original resolution without extracting multi-scale features adversely affects its effectiveness, which does not align with our model's design that leverages cross-scale interactions. When incorporating VMamba as the visual encoder, it achieves comparable runtime efficiency to PVT-v2 but exhibits a modest performance decline under the MS3 setting.

% as shown in Tab.~\ref{Ablation study of scan order}. Our findings indicate that employing 8 scanning directions yields optimal performance. This is attributed to Mamba's directional and sequential sensitivity, which restricts the ability of individual points to absorb information from subsequent features. By expanding the scanning orientations, we facilitate a more holistic perception and understanding of temporal-spatial information for each point, enabling the extraction of more comprehensive and enriched features.

% To more intuitively explain the role of the Mamba modules, we visualize the feature maps extracted from the visual backbone alongside those that have been integrated using the CIP module. As illustrated in Fig.~\ref{fig:feature visualization} (c) and (d), we employ the principal component analysis (PCA)~\cite{shlens2014tutorial} technique to extract the principal channel of the feature maps and combine them with the original images with the scaling factor of 0.7, with highlighted regions indicating the locations focused by the model.
% It can be observed that when audio features are not integrated, the model primarily focuses on the background regions. However, once the audio-to-vision feature aggregation is performed, the model’s attention significantly shifts to zones containing the sound-emitting objects. The result qualitatively demonstrates the effectiveness of the CIP module in enhancing the cross-modal capabilities of the Mamba.

\section{Conclusion}
We present AVS-Mamba, a framework based on the Mamba architecture and tailored for audio-visual segmentation. AVS-Mamba significantly refines the original Mamba design, capitalizing on its proficient handling of long sequences to achieve cross-scale temporal consistency and enhanced audio-visual modality comprehension. The framework initiates with the development of the Multi-scale Temporal Encoder, designed for extracting features across scales within and between frames. We then present the Modality Aggregation Decoder, featuring a Vision-to-Audio Fusion Block that seamlessly integrates visual data from diverse spatiotemporal contexts into the audio stream. Additionally, the Contextual Integration Pyramid is developed to systematically gather fine-grained features across different frames and modalities. Our extensive experiments confirm that AVS-Mamba delivers state-of-the-art results on the AVSBench datasets. We anticipate that our work will pave the way for further explorations of the Mamba architecture in addressing audio-visual learning and perception challenges.

% \section*{Acknowledgments}
% This should be a simple paragraph before the References to thank those individuals and institutions who have supported your work on this article.


\bibliographystyle{IEEEtran}
\bibliography{ieeetran}

\begin{thebibliography}{10}
\providecommand{\url}[1]{#1}
\csname url@samestyle\endcsname
\providecommand{\newblock}{\relax}
\providecommand{\bibinfo}[2]{#2}
\providecommand{\BIBentrySTDinterwordspacing}{\spaceskip=0pt\relax}
\providecommand{\BIBentryALTinterwordstretchfactor}{4}
\providecommand{\BIBentryALTinterwordspacing}{\spaceskip=\fontdimen2\font plus
\BIBentryALTinterwordstretchfactor\fontdimen3\font minus \fontdimen4\font\relax}
\providecommand{\BIBforeignlanguage}[2]{{%
\expandafter\ifx\csname l@#1\endcsname\relax
\typeout{** WARNING: IEEEtran.bst: No hyphenation pattern has been}%
\typeout{** loaded for the language `#1'. Using the pattern for}%
\typeout{** the default language instead.}%
\else
\language=\csname l@#1\endcsname
\fi
#2}}
\providecommand{\BIBdecl}{\relax}
\BIBdecl

\bibitem{zhou2022audio}
J.~Zhou, J.~Wang, J.~Zhang, W.~Sun, J.~Zhang, S.~Birchfield, D.~Guo, L.~Kong, M.~Wang, and Y.~Zhong, ``Audio--visual segmentation,'' in \emph{Eur. Conf. Comput. Vis.}\hskip 1em plus 0.5em minus 0.4em\relax Springer, 2022, pp. 386--403.

\bibitem{zhou2023audio}
J.~Zhou, X.~Shen, J.~Wang, J.~Zhang, W.~Sun, J.~Zhang, S.~Birchfield, D.~Guo, L.~Kong, M.~Wang \emph{et~al.}, ``Audio-visual segmentation with semantics,'' \emph{arXiv preprint arXiv:2301.13190}, 2023.

\bibitem{hao2023improving}
D.~Hao, Y.~Mao, B.~He, X.~Han, Y.~Dai, and Y.~Zhong, ``Improving audio-visual segmentation with bidirectional generation,'' \emph{arXiv preprint arXiv:2308.08288}, 2023.

\bibitem{gao2023avsegformer}
S.~Gao, Z.~Chen, G.~Chen, W.~Wang, and T.~Lu, ``Avsegformer: Audio-visual segmentation with transformer,'' \emph{arXiv preprint arXiv:2307.01146}, 2023.

\bibitem{huang2023discovering}
S.~Huang, H.~Li, Y.~Wang, H.~Zhu, J.~Dai, J.~Han, W.~Rong, and S.~Liu, ``Discovering sounding objects by audio queries for audio visual segmentation,'' \emph{arXiv preprint arXiv:2309.09501}, 2023.

\bibitem{li2023catr}
K.~Li, Z.~Yang, L.~Chen, Y.~Yang, and J.~Xiao, ``Catr: Combinatorial-dependence audio-queried transformer for audio-visual video segmentation,'' in \emph{ACM Int. Conf. Multimedia}, 2023, pp. 1485--1494.

\bibitem{chen2024bootstrapping}
T.~Chen, Z.~Tan, T.~Gong, Q.~Chu, Y.~Wu, B.~Liu, L.~Lu, J.~Ye, and N.~Yu, ``Bootstrapping audio-visual segmentation by strengthening audio cues,'' \emph{arXiv preprint arXiv:2402.02327}, 2024.

\bibitem{gu2023mamba}
A.~Gu and T.~Dao, ``Mamba: Linear-time sequence modeling with selective state spaces,'' \emph{arXiv preprint arXiv:2312.00752}, 2023.

\bibitem{zhu2024vision}
L.~Zhu, B.~Liao, Q.~Zhang, X.~Wang, W.~Liu, and X.~Wang, ``Vision mamba: Efficient visual representation learning with bidirectional state space model,'' \emph{arXiv preprint arXiv:2401.09417}, 2024.

\bibitem{liu2024vmamba}
Y.~Liu, Y.~Tian, Y.~Zhao, H.~Yu, L.~Xie, Y.~Wang, Q.~Ye, and Y.~Liu, ``Vmamba: Visual state space model,'' \emph{arXiv preprint arXiv:2401.10166}, 2024.

\bibitem{lin2017feature}
T.-Y. Lin, P.~Doll{\'a}r, R.~Girshick, K.~He, B.~Hariharan, and S.~Belongie, ``Feature pyramid networks for object detection,'' in \emph{IEEE Conf. Comput. Vis. Pattern Recognit.}, 2017, pp. 2117--2125.

\bibitem{xiong2022look}
J.~Xiong, Y.~Zhou, P.~Zhang, L.~Xie, W.~Huang, and Y.~Zha, ``Look\&listen: Multi-modal correlation learning for active speaker detection and speech enhancement,'' \emph{IEEE Trans. Multimedia}, vol.~25, pp. 5800--5812, 2022.

\bibitem{wang2020robust}
W.~Wang, C.~Xing, D.~Wang, X.~Chen, and F.~Sun, ``A robust audio-visual speech enhancement model,'' in \emph{Int. Conf. Acous., Speech, Sign. Process.}\hskip 1em plus 0.5em minus 0.4em\relax IEEE, 2020, pp. 7529--7533.

\bibitem{xue2021audio}
C.~Xue, X.~Zhong, M.~Cai, H.~Chen, and W.~Wang, ``Audio-visual event localization by learning spatial and semantic co-attention,'' \emph{IEEE Trans. Multimedia}, vol.~25, pp. 418--429, 2021.

\bibitem{jiang2023leveraging}
Y.~Jiang, J.~Yin, and Y.~Dang, ``Leveraging the video-level semantic consistency of event for audio-visual event localization,'' \emph{IEEE Trans. Multimedia}, 2023.

\bibitem{liu2022dense}
S.~Liu, W.~Quan, C.~Wang, Y.~Liu, B.~Liu, and D.-M. Yan, ``Dense modality interaction network for audio-visual event localization,'' \emph{IEEE Trans. Multimedia}, vol.~25, pp. 2734--2748, 2022.

\bibitem{feng2023css}
F.~Feng, Y.~Ming, N.~Hu, H.~Yu, and Y.~Liu, ``Css-net: A consistent segment selection network for audio-visual event localization,'' \emph{IEEE Trans. Multimedia}, 2023.

\bibitem{liu2022visual}
X.~Liu, R.~Qian, H.~Zhou, D.~Hu, W.~Lin, Z.~Liu, B.~Zhou, and X.~Zhou, ``Visual sound localization in the wild by cross-modal interference erasing,'' in \emph{AAAI Conf. Arti. Intell.}, vol.~36, no.~2, 2022, pp. 1801--1809.

\bibitem{mo2023audio}
S.~Mo and Y.~Tian, ``Audio-visual grouping network for sound localization from mixtures,'' in \emph{IEEE Conf. Comput. Vis. Pattern Recognit.}, 2023, pp. 10\,565--10\,574.

\bibitem{fu2023multimodal}
J.~Fu, J.~Gao, B.-K. Bao, and C.~Xu, ``Multimodal imbalance-aware gradient modulation for weakly-supervised audio-visual video parsing,'' \emph{IEEE Trans. Cir. Syst. Video Tech.}, 2023.

\bibitem{xian2023vita}
K.~Xian, J.~Peng, Z.~Cao, J.~Zhang, and G.~Lin, ``Vita: Video transformer adaptor for robust video depth estimation,'' \emph{IEEE Trans. Multimedia}, 2023.

\bibitem{li2024adaptive}
Q.~Li, G.~Zu, H.~Xu, J.~Kong, Y.~Zhang, and J.~Wang, ``An adaptive dual selective transformer for temporal action localization,'' \emph{IEEE Trans. Multimedia}, 2024.

\bibitem{zhang2023end}
Y.~Zhang, Y.~Pan, T.~Yao, R.~Huang, T.~Mei, and C.-W. Chen, ``End-to-end video scene graph generation with temporal propagation transformer,'' \emph{IEEE Trans. Multimedia}, vol.~26, pp. 1613--1625, 2023.

\bibitem{zhuge2024learning}
Y.~Zhuge, H.~Gu, L.~Zhang, J.~Qi, and H.~Lu, ``Learning motion and temporal cues for unsupervised video object segmentation,'' \emph{IEEE Trans. Neural Netw. Learn. Syst.}, 2024.

\bibitem{dosovitskiy2020image}
A.~Dosovitskiy, L.~Beyer, A.~Kolesnikov, D.~Weissenborn, X.~Zhai, T.~Unterthiner, M.~Dehghani, M.~Minderer, G.~Heigold, S.~Gelly \emph{et~al.}, ``An image is worth 16x16 words: Transformers for image recognition at scale,'' \emph{arXiv preprint arXiv:2010.11929}, 2020.

\bibitem{touvron2021training}
H.~Touvron, M.~Cord, M.~Douze, F.~Massa, A.~Sablayrolles, and H.~J{\'e}gou, ``Training data-efficient image transformers \& distillation through attention,'' in \emph{Int. Conf. Mach. Learn.}, 2021, pp. 10\,347--10\,357.

\bibitem{liu2021swin}
Z.~Liu, Y.~Lin, Y.~Cao, H.~Hu, Y.~Wei, Z.~Zhang, S.~Lin, and B.~Guo, ``Swin transformer: Hierarchical vision transformer using shifted windows,'' in \emph{Int. Conf. Comput. Vis.}, 2021, pp. 10\,012--10\,022.

\bibitem{yang2024vivim}
Y.~Yang, Z.~Xing, and L.~Zhu, ``Vivim: a video vision mamba for medical video object segmentation,'' \emph{arXiv preprint arXiv:2401.14168}, 2024.

\bibitem{yue2024medmamba}
Y.~Yue and Z.~Li, ``Medmamba: Vision mamba for medical image classification,'' \emph{arXiv preprint arXiv:2403.03849}, 2024.

\bibitem{zhang2024vm}
M.~Zhang, Y.~Yu, L.~Gu, T.~Lin, and X.~Tao, ``Vm-unet-v2 rethinking vision mamba unet for medical image segmentation,'' \emph{arXiv preprint arXiv:2403.09157}, 2024.

\bibitem{ye2024p}
Z.~Ye and T.~Chen, ``P-mamba: Marrying perona malik diffusion with mamba for efficient pediatric echocardiographic left ventricular segmentation,'' \emph{arXiv preprint arXiv:2402.08506}, 2024.

\bibitem{ma2024rs}
X.~Ma, X.~Zhang, and M.-O. Pun, ``Rs 3 mamba: Visual state space model for remote sensing image semantic segmentation,'' \emph{IEEE Geos. Remote Sensing Letters}, 2024.

\bibitem{chen2024rsmamba}
K.~Chen, B.~Chen, C.~Liu, W.~Li, Z.~Zou, and Z.~Shi, ``Rsmamba: Remote sensing image classification with state space model,'' \emph{IEEE Geos. Remote Sensing Letters}, 2024.

\bibitem{liu2024rscama}
C.~Liu, K.~Chen, B.~Chen, H.~Zhang, Z.~Zou, and Z.~Shi, ``Rscama: Remote sensing image change captioning with state space model,'' \emph{IEEE Geos. Remote Sensing Letters}, 2024.

\bibitem{chen2024changemamba}
H.~Chen, J.~Song, C.~Han, J.~Xia, and N.~Yokoya, ``Changemamba: Remote sensing change detection with spatio-temporal state space model,'' \emph{arXiv preprint arXiv:2404.03425}, 2024.

\bibitem{guo2024mambair}
H.~Guo, J.~Li, T.~Dai, Z.~Ouyang, X.~Ren, and S.-T. Xia, ``Mambair: A simple baseline for image restoration with state-space model,'' \emph{arXiv preprint arXiv:2402.15648}, 2024.

\bibitem{zheng2024u}
Z.~Zheng and C.~Wu, ``U-shaped vision mamba for single image dehazing,'' \emph{arXiv preprint arXiv:2402.04139}, 2024.

\bibitem{shi2024vmambair}
Y.~Shi, B.~Xia, X.~Jin, X.~Wang, T.~Zhao, X.~Xia, X.~Xiao, and W.~Yang, ``Vmambair: Visual state space model for image restoration,'' \emph{arXiv preprint arXiv:2403.11423}, 2024.

\bibitem{qiao2024vlmamba}
Y.~Qiao, Z.~Yu, L.~Guo, S.~Chen, Z.~Zhao, M.~Sun, Q.~Wu, and J.~Liu, ``Vl-mamba: Exploring state space models for multimodal learning,'' 2024.

\bibitem{zhao2024cobra}
H.~Zhao, M.~Zhang, W.~Zhao, P.~Ding, S.~Huang, and D.~Wang, ``Cobra: Extending mamba to multi-modal large language model for efficient inference,'' 2024.

\bibitem{yang2024remamber}
Y.~Yang, C.~Ma, J.~Yao, Z.~Zhong, Y.~Zhang, and Y.~Wang, ``Remamber: Referring image segmentation with mamba twister,'' \emph{arXiv preprint arXiv:2403.17839}, 2024.

\bibitem{liu2024robomamba}
J.~Liu, M.~Liu, Z.~Wang, L.~Lee, K.~Zhou, P.~An, S.~Yang, R.~Zhang, Y.~Guo, and S.~Zhang, ``Robomamba: Multimodal state space model for efficient robot reasoning and manipulation,'' \emph{arXiv preprint arXiv:2406.04339}, 2024.

\bibitem{chen2024video}
G.~Chen, Y.~Huang, J.~Xu, B.~Pei, Z.~Chen, Z.~Li, J.~Wang, K.~Li, T.~Lu, and L.~Wang, ``Video mamba suite: State space model as a versatile alternative for video understanding,'' \emph{arXiv preprint arXiv:2403.09626}, 2024.

\bibitem{li2024videomamba}
K.~Li, X.~Li, Y.~Wang, Y.~He, Y.~Wang, L.~Wang, and Y.~Qiao, ``Videomamba: State space model for efficient video understanding,'' \emph{arXiv preprint arXiv:2403.06977}, 2024.

\bibitem{lu2024videomambapro}
H.~Lu, A.~A. Salah, and R.~Poppe, ``Videomambapro: A leap forward for mamba in video understanding,'' \emph{arXiv preprint arXiv:2406.19006}, 2024.

\bibitem{wang2024mamba}
Z.~Wang, F.~Kong, S.~Feng, M.~Wang, H.~Zhao, D.~Wang, and Y.~Zhang, ``Is mamba effective for time series forecasting?'' \emph{arXiv preprint arXiv:2403.11144}, 2024.

\bibitem{chen2024mim}
T.~Chen, Z.~Tan, T.~Gong, Q.~Chu, Y.~Wu, B.~Liu, J.~Ye, and N.~Yu, ``Mim-istd: Mamba-in-mamba for efficient infrared small target detection,'' \emph{arXiv preprint arXiv:2403.02148}, 2024.

\bibitem{pei2024efficientvmamba}
X.~Pei, T.~Huang, and C.~Xu, ``Efficientvmamba: Atrous selective scan for light weight visual mamba,'' \emph{arXiv preprint arXiv:2403.09977}, 2024.

\bibitem{huang2024localmamba}
T.~Huang, X.~Pei, S.~You, F.~Wang, C.~Qian, and C.~Xu, ``Localmamba: Visual state space model with windowed selective scan,'' \emph{arXiv preprint arXiv:2403.09338}, 2024.

\bibitem{yang2024plainmamba}
C.~Yang, Z.~Chen, M.~Espinosa, L.~Ericsson, Z.~Wang, J.~Liu, and E.~J. Crowley, ``Plainmamba: Improving non-hierarchical mamba in visual recognition,'' \emph{arXiv preprint arXiv:2403.17695}, 2024.

\bibitem{shi2024multi}
Y.~Shi, M.~Dong, and C.~Xu, ``Multi-scale vmamba: Hierarchy in hierarchy visual state space model,'' \emph{arXiv preprint arXiv:2405.14174}, 2024.

\bibitem{huangav}
Z.~Huang, J.~Li, W.~Zhao, Y.~Guo, and Y.~Tian, ``Av-mamba: Cross-modality selective state space models for audio-visual question answering.''

\bibitem{li2022learning}
G.~Li, Y.~Wei, Y.~Tian, C.~Xu, J.-R. Wen, and D.~Hu, ``Learning to answer questions in dynamic audio-visual scenarios,'' in \emph{Proceedings of the IEEE/CVF Conference on Computer Vision and Pattern Recognition}, 2022, pp. 19\,108--19\,118.

\bibitem{li2024selm}
J.~Li, S.~Yu, Y.~Wang, L.~Wang, and H.~Lu, ``Selm: Selective mechanism based audio-visual segmentation,'' in \emph{ACM Multimedia 2024}.

\bibitem{kalman1960new}
R.~E. Kalman, ``A new approach to linear filtering and prediction problems,'' 1960.

\bibitem{zhang2023hivit}
X.~Zhang, Y.~Tian, L.~Xie, W.~Huang, Q.~Dai, Q.~Ye, and Q.~Tian, ``Hivit: A simpler and more efficient design of hierarchical vision transformer,'' in \emph{Int. Conf. Learn. Represent.}, 2023.

\bibitem{chen2024res}
C.-S. Chen, G.-Y. Chen, D.~Zhou, D.~Jiang, and D.-S. Chen, ``Res-vmamba: Fine-grained food category visual classification using selective state space models with deep residual learning,'' \emph{arXiv preprint arXiv:2402.15761}, 2024.

\bibitem{zhu2020deformable}
X.~Zhu, W.~Su, L.~Lu, B.~Li, X.~Wang, and J.~Dai, ``Deformable detr: Deformable transformers for end-to-end object detection,'' \emph{arXiv preprint arXiv:2010.04159}, 2020.

\bibitem{ba2016layer}
J.~L. Ba, J.~R. Kiros, and G.~E. Hinton, ``Layer normalization,'' \emph{arXiv preprint arXiv:1607.06450}, 2016.

\bibitem{xu2024each}
S.~Xu, S.~Wei, T.~Ruan, L.~Liao, and Y.~Zhao, ``Each perform its functions: Task decomposition and feature assignment for audio-visual segmentation,'' \emph{IEEE Trans. Multimedia}, 2024.

\bibitem{mao2023contrastive}
Y.~Mao, J.~Zhang, M.~Xiang, Y.~Lv, Y.~Zhong, and Y.~Dai, ``Contrastive conditional latent diffusion for audio-visual segmentation,'' \emph{arXiv preprint arXiv:2307.16579}, 2023.

\bibitem{liu2024bavs}
C.~Liu, P.~Li, H.~Zhang, L.~Li, Z.~Huang, D.~Wang, and X.~Yu, ``Bavs: bootstrapping audio-visual segmentation by integrating foundation knowledge,'' \emph{IEEE Trans. Multimedia}, 2024.

\bibitem{mao2023multimodal}
Y.~Mao, J.~Zhang, M.~Xiang, Y.~Zhong, and Y.~Dai, ``Multimodal variational auto-encoder based audio-visual segmentation,'' in \emph{Int. Conf. Comput. Vis.}, 2023, pp. 954--965.

\bibitem{he2016deep}
K.~He, X.~Zhang, S.~Ren, and J.~Sun, ``Deep residual learning for image recognition,'' in \emph{IEEE Conf. Comput. Vis. Pattern Recognit.}, 2016, pp. 770--778.

\bibitem{lin2014microsoft}
T.-Y. Lin, M.~Maire, S.~Belongie, J.~Hays, P.~Perona, D.~Ramanan, P.~Doll{\'a}r, and C.~L. Zitnick, ``Microsoft coco: Common objects in context,'' in \emph{Eur. Conf. Comput. Vis.}\hskip 1em plus 0.5em minus 0.4em\relax Springer, 2014, pp. 740--755.

\bibitem{wang2022pvt}
W.~Wang, E.~Xie, X.~Li, D.-P. Fan, K.~Song, D.~Liang, T.~Lu, P.~Luo, and L.~Shao, ``Pvt v2: Improved baselines with pyramid vision transformer,'' \emph{Comput. Visual Media}, vol.~8, no.~3, pp. 415--424, 2022.

\bibitem{russakovsky2015imagenet}
O.~Russakovsky, J.~Deng, H.~Su, J.~Krause, S.~Satheesh, S.~Ma, Z.~Huang, A.~Karpathy, A.~Khosla, M.~Bernstein \emph{et~al.}, ``Imagenet large scale visual recognition challenge,'' \emph{Int. J. Comput. Vis.}, vol. 115, pp. 211--252, 2015.

\bibitem{arandjelovic2017look}
R.~Arandjelovic and A.~Zisserman, ``Look, listen and learn,'' in \emph{Int. Conf. Comput. Vis.}, 2017, pp. 609--617.

\bibitem{gemmeke2017audio}
J.~F. Gemmeke, D.~P. Ellis, D.~Freedman, A.~Jansen, W.~Lawrence, R.~C. Moore, M.~Plakal, and M.~Ritter, ``Audio set: An ontology and human-labeled dataset for audio events,'' in \emph{Int. Conf. Acous., Speech, Sign. Process.}\hskip 1em plus 0.5em minus 0.4em\relax IEEE, 2017, pp. 776--780.

\bibitem{shlens2014tutorial}
J.~Shlens, ``A tutorial on principal component analysis,'' \emph{arXiv preprint arXiv:1404.1100}, 2014.

\end{thebibliography}

\end{document}